\documentclass[pdflatex,sn-mathphys-num]{sn-jnl}% Math and Physical Sciences Numbered Reference Style
\usepackage{subcaption}
\usepackage{graphicx}%
\usepackage{multirow}%
\usepackage{amsmath,amssymb,amsfonts}%
\usepackage{amsthm}%
\usepackage{mathrsfs}%
\usepackage[title]{appendix}%
\usepackage{xcolor}%
\usepackage{textcomp}%
\usepackage{manyfoot}%
\usepackage{booktabs}%
\usepackage{algorithm}%
\usepackage{algorithmicx}%
\usepackage{algpseudocode}%
\usepackage{listings}%
\usepackage{pgfplots}      % 统计图表
\usepackage{pifont}        % \ding命令(勾叉符号)
\usepackage{makecell}      % 表头换行
\def \Ours {DoctorAgent-RL}
\usepackage{orcidlink}
\newcommand{\cmark}{\textcolor{green!60!black}{\ding{51}}}
\newcommand{\xmark}{\textcolor{red}{\ding{55}}}
\theoremstyle{thmstyleone}%
\theoremstyle{thmstyletwo}%

\theoremstyle{thmstylethree}%

\begin{document}

\title{Real-World Doctor Agent with Proactive Consultation through Multi-Agent Reinforcement Learning}


\author[1,2,3]{\fnm{Yichun} \sur{Feng}\orcidlink{0009-0007-4511-4713}}\email{fengyichun22@mails.ucas.ac.cn}\equalcont{These authors contributed equally to this work.}

\author[4]{\fnm{Jiawei} \sur{Wang}\orcidlink{0009-0005-0575-2695}}\email{wangjiawei@mail.ustc.edu.cn}
\equalcont{These authors contributed equally to this work.}

\author[3]{\fnm{Lu} \sur{Zhou}\orcidlink{0009-0003-6538-0649}}\email{zhou\_lu@gzlab.ac.cn}
\author[5]{\fnm{Yikai} \sur{Zheng}\orcidlink{0009-0004-8709-4536}}\email{zhengyk23@mail2.sysu.edu.cn}

\author*[1,2,9]{\fnm{Zhen} \sur{Lei}\orcidlink{0000-0002-0791-189X}}\email{zhen.lei@ia.ac.cn}
\author*[3,6,7,8]{\fnm{Yixue} \sur{Li}\orcidlink{0000-0002-1198-7176}}\email{li\_yixue@gzlab.ac.cn}

\affil[1]{\orgdiv{School of Advanced Interdisciplinary Sciences}, \orgname{University of Chinese Academy of Sciences}, \orgaddress{\street{380, Huaibei Town, Huairou District}, \city{Beijing}, \postcode{100049}, \state{Beijing}, \country{China}}}

\affil[2]{\orgdiv{Institute of Automation}, \orgname{Chinese Academy of Sciences}, \orgaddress{\street{No. 95, Zhongguancun East Road}, \city{Beijing}, \postcode{100190}, \country{China}}}

\affil[3]{\orgname{Guangzhou National Laboratory}, \orgaddress{\street{No. 9 XingDaoHuanBei Road, Guangzhou International Bio Island}, \city{Guangzhou}, \postcode{510005}, \state{Guangdong}, \country{China}}}

\affil[4]{\orgdiv{Department of EEIS}, \orgname{University of Science and Technology of China}, \orgaddress{\street{No. 443, Huangshan Road, Shushan District}, \city{Hefei}, \postcode{230026}, \state{Anhui}, \country{China}}}
\affil[5]{\orgdiv{School of Intelligent Systems Engineering}, \orgname{Sun Yat-sen University}, \orgaddress{\street{No. 66, Gongchang Road, Guangming District}, \city{Shenzhen}, \postcode{518107}, \state{Guangdong}, \country{China}}}
\affil[6]{\orgdiv{GMU-GIBH Joint School of Life Sciences}, \orgname{The Guangdong-Hong Kong-Macau Joint Laboratory for Cell Fate Regulation and Diseases}, \orgname{Guangzhou Medical University}, \orgaddress{\city{Guangzhou}, \postcode{511436}, \country{China}}}

\affil[7]{\orgdiv{School of Life Sciences and Biotechnology}, \orgname{Shanghai Jiao Tong University}, \orgaddress{\city{Shanghai}, \postcode{200240}, \country{China}}}

\affil[8]{\orgdiv{Shanghai Institute of Nutrition and Health}, \orgname{Chinese Academy of Sciences}, \orgaddress{\city{Shanghai}, \postcode{200030}, \country{China}}}

% \affil[9]{\orgdiv{Bioland Laboratory}, \orgaddress{\city{Guangzhou}, \postcode{510005}, \country{China}}}

\affil[9]{\orgdiv{Centre for Artificial Intelligence and Robotics, Hong Kong Institute of Science \& Innovation}, \orgname{Chinese Academy of Sciences}, \orgaddress{\city{Hong Kong}, \country{China}}}

\abstract{Large language models (LLMs) struggle in real-world clinical consultations. Single-turn consultation systems require patients to describe all symptoms at once, which often leads to unclear complaints and vague diagnoses. Traditional dialogue models, constrained by static supervised learning, are limited to superficially imitating existing dialogue patterns and lack the ability to actively construct understanding in dynamic interactions, thus failing to achieve genuine clinical reasoning.
To address these challenges, we propose DoctorAgent-RL, a reinforcement learning (RL)-based multi-agent collaborative framework, and train a doctor agent on Qwen2.5-7B-Instruct using this framework. Within this framework, a medical consultation is modeled as a dynamic decision-making process under uncertainty. The core intelligence of the doctor agent is shifted from knowing the answer to learning and mastering a questioning methodology aimed at achieving an optimal diagnosis. Through strategic questioning, it guides the progressive emergence of key patient information in multi-turn dialogues. To support this high-fidelity simulation of the real diagnostic process, we constructed MTMedDialog, a novel English multi-turn medical consultation dataset designed for dynamic, interactive training.
To validate its real-world effectiveness, rigorous evaluations including blinded human assessments and trials with real patients were conducted. DoctorAgent-RL outperformed frontier models and achieved a 70\% exact diagnostic match rate, confirming its potential as a collaborative tool. By handling initial screenings, it can free clinicians to focus on complex cases, thereby addressing critical issues like physician shortages and misdiagnosis risks while alleviating the strain on healthcare resources.}




\maketitle

\section{Introduction}\label{sec1}

Large language models (LLMs) like ChatGPT \cite{achiam2023gpt}, LLaMA \cite{touvron2023llama}, and ChatGLM \cite{glm2024chatglm} show significant potential for assisting in clinical decision-making \cite{kopka2025accuracy}. This capability is urgently needed in the context of a global doctor shortage and high-pressure healthcare environments \cite{agyeman2023prioritising}. In these settings, time pressure on clinicians often leads to rapid diagnoses based on incomplete information, increasing misdiagnosis risks \cite{alqahtani2016does}, while routine consultations consume resources that could be allocated to more critical cases \cite{arndt2024more}. Therefore, developing an intelligent system to improve primary care efficiency holds significant practical importance for optimizing resource allocation and alleviating healthcare strain \cite{al2023review}.

The limitations of current systems are evident on multiple levels. First, single-turn question-answering systems, such as MedAlpaca \cite{han2023medalpaca} and BioMistral \cite{labrak2024biomistral}, require patients to provide a comprehensive description of their condition all at once \cite{wu2025towards}. This contradicts the clinical reality of complex situations where chief complaints are unclear and information is vague \cite{laban2025llms}, often leading to overly broad or even risky diagnostic suggestions \cite{auerbach2024diagnostic}. Second, even traditional models that support multi-turn dialogue, like Bianque \cite{chen2023bianque} and DialoGPT \cite{zhang2019dialogpt}, rely heavily on static supervised learning, which amounts to a superficial imitation of existing dialogue transcripts \cite{liu2025interactive}. These models lack a dynamic decision-making mechanism and cannot intelligently adjust their questioning strategies based on the real-time progress of the conversation to probe for key information. Even systems that integrate knowledge graphs \cite{wang2023huatuo, zhang2024ultramedical, feng2025knowledge} fundamentally rely on predefined dialogue paths, failing to achieve genuine clinical reasoning. This fundamentally solidifies the role of LLM as a passive knowledge reproducer, whose intelligence remains limited to reproducing existing knowledge rather than actively exploring the unknown.

In recent years, research has begun to explore more dynamic approaches. For instance, systems like AMIE \cite{tu2025towards} optimize the diagnostic dialogue itself through self-play simulations, while multi-agent frameworks such as Agent Hospital \cite{li2024agent}, AI Hospital \cite{fan2024ai}, and MAC \cite{chen2025enhancing} simulate complex hospital environments or multi-role collaboration protocols. Additionally, EvoMDT \cite{liu2026evomdt} introduces a self-evolving multi-agent system that automatically optimizes prompts and retrieval scopes based on feedback. However, these methods fundamentally rely on prompt-level optimizations and the synthesis of static information. They still lack the capability to optimize proactive inquiry strategies when facing the inherent uncertainty of patient-led interactions. Concurrently, the application of reinforcement learning (RL) in the medical field is transitioning toward such dynamic paradigms. MedVLM-R1 \cite{pan2025medvlm} and Med-R1 \cite{lai2025med} employ the Group Relative Policy Optimization (GRPO) framework \cite{shao2024deepseekmath} to enhance diagnostic accuracy, while HuatuoGPT-o1 \cite{chen2024huatuogpt} improves clinical reasoning through verifiable question generation. MedRIA \cite{zou2024ai} and PPME \cite{sun2025improving} further explore inquiry efficiency and experience replay. Despite these efforts, existing approaches lack robust frameworks for modeling dynamic doctor-patient interactions and validating performance in real clinical settings with high complexity and uncertainty.

\begin{figure*}[]
    \centering
    \includegraphics[width=1.0\textwidth]{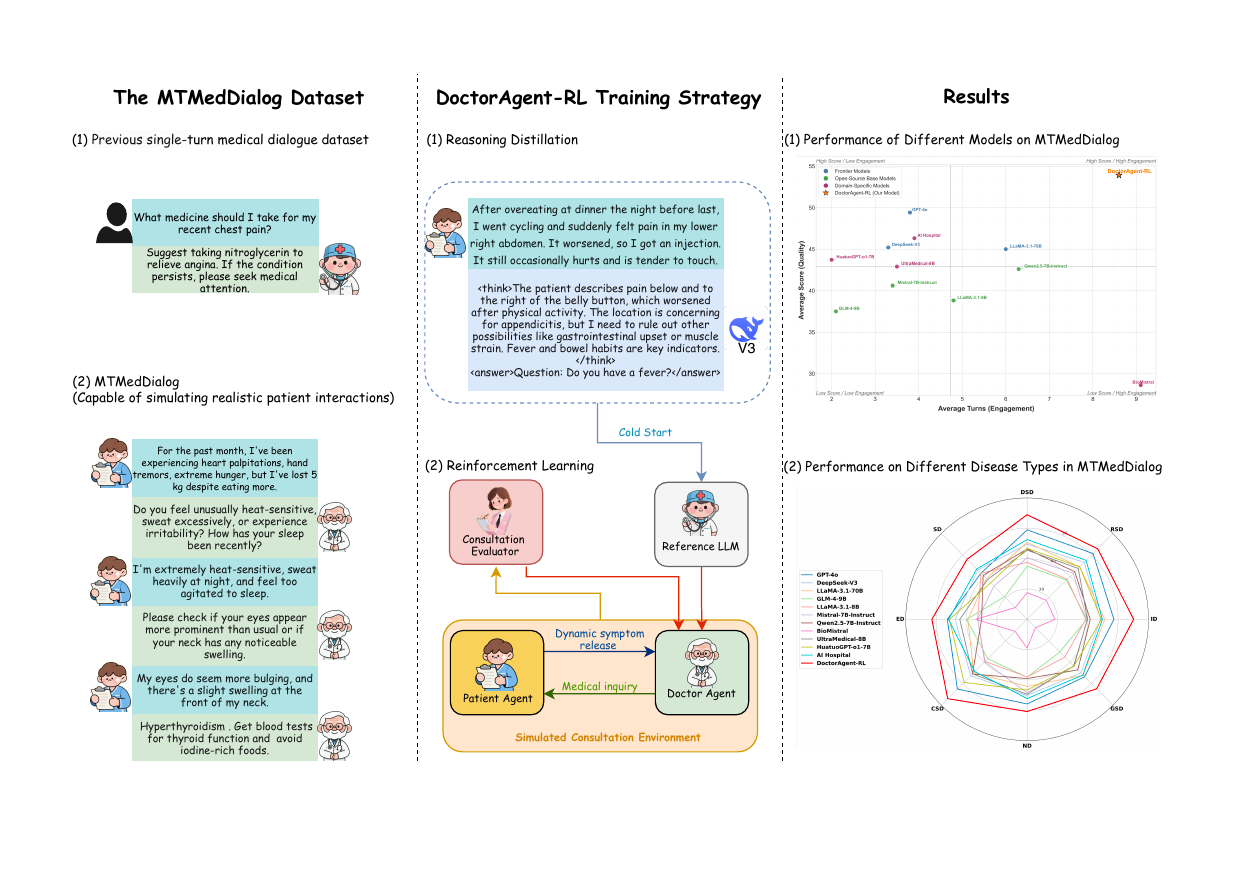} 
    \caption{Overview of DoctorAgent-RL Framework and Performance. \textbf{The MTMedDialog Dataset:} A high-fidelity multi-turn medical dialogue dataset supporting dynamic patient simulation, where realistic patient responses are generated based on comprehensive medical profiles. \textbf{DoctorAgent-RL Training Strategy:} A multi-agent reinforcement learning framework combining reasoning distillation and policy optimization, where the doctor agent learns strategic questioning through interaction with the patient agent under guidance from a consultation evaluator. \textbf{Results:} \Ours{} achieves state-of-the-art on the MTMedDialog task.} % 标题
    \label{fig:first} 
\end{figure*}

To address these challenges, we propose \Ours{}—a multi-agent collaborative reinforcement learning framework that reformulates clinical reasoning as a Markov Decision Process (MDP), as shown in Figure~\ref{fig:first}. Within this framework: (1) A high-fidelity patient agent based on LLMs, which generates pathologically consistent responses while mimicking the diversity of real-world communication; (2) A doctor agent initialized by cloning medical actions from real consultation records and refined through RL to master effective questioning strategies; and (3) A consultation evaluator that provides multi-dimensional rewards based on diagnostic accuracy, patient information responsiveness, and the standardization of questions. Through this framework, we train the \Ours{} model based on Qwen2.5-7B-Instruct \cite{xu2025qwen2}, which redefines the diagnostic process through reward-based strategic policy optimization: the doctor agent continuously optimizes its questioning strategy, obtaining immediate feedback through multi-turn interactions with the patient agent, and dynamically adjusting the information-gathering path based on comprehensive rewards from the consultation evaluator. Here, the core intelligence of the doctor agent no longer lies in knowing the answer, but rather in learning and mastering a questioning methodology aimed at achieving optimal diagnosis through strategic questioning that guides the gradual emergence of key patient information throughout multi-turn dialogues. This end-to-end RL fine-tuning mechanism enables LLMs to autonomously evolve interaction strategies that align with clinical reasoning logic, rather than simply imitating surface patterns in existing dialogue data.

While existing datasets provide valuable examples of multi-turn medical conversations, they consist of static transcripts. This limits their use for training agents that need to learn adaptive questioning strategies through interactive simulation. To address this gap, we constructed MTMedDialog, the first English dataset specifically designed to support dynamic, stateful patient simulations. Each dialogue is grounded in a comprehensive patient profile, allowing for realistic, turn-by-turn information revelation based on the clinician agent's diagnostic reasoning process. This high-fidelity simulation of the actual diagnostic process is an essential prerequisite for training a knowledge constructor, as it enables the agent to learn and validate its questioning strategies in a near-realistic environment.

This manuscript presents a substantial extension of our conference work \cite{feng2025Doctor}. While that study established the technical feasibility of enabling AI doctors to conduct proactive, strategic questioning through reinforcement learning, yet a critical question remained unvalidated: can LLMs trained through simulation maintain diagnostic accuracy and adaptive reasoning when confronted with real patients' inherent unpredictability? Here, we provide definitive answers across three dimensions—clinical efficacy, generalization capability, and systematic completeness. Prospective evaluation with actual patients achieved 70\% exact diagnostic match rate, blinded human assessment confirmed superior comprehensive clinical performance, cross-dataset evaluation on HealthBench \cite{arora2025healthbench} validated transferable reasoning patterns, and systematic analyses further established the preservation of general-purpose capabilities alongside patient agent simulation fidelity. These findings establish that strategically trained AI agents can function as reliable clinical collaborators rather than brittle pattern matchers in unpredictable real-world settings, opening a viable pathway to address the global healthcare workforce crisis. DoctorAgent-RL can handle initial consultations and routine triage, freeing human physicians to focus on complex cases requiring nuanced judgment, thereby improving diagnostic quality and resource allocation efficiency across healthcare systems worldwide.

\section{Results}\label{sec2}
\subsection{DoctorAgent-RL Framework Overview}
\label{sec:framework}

\begin{figure*}[!htbp]
    \centering
    
    % ============ 第1行: 图(a) 占满整行 ============
    \begin{subfigure}[t]{\textwidth}
        \centering
        \includegraphics[width=\textwidth]{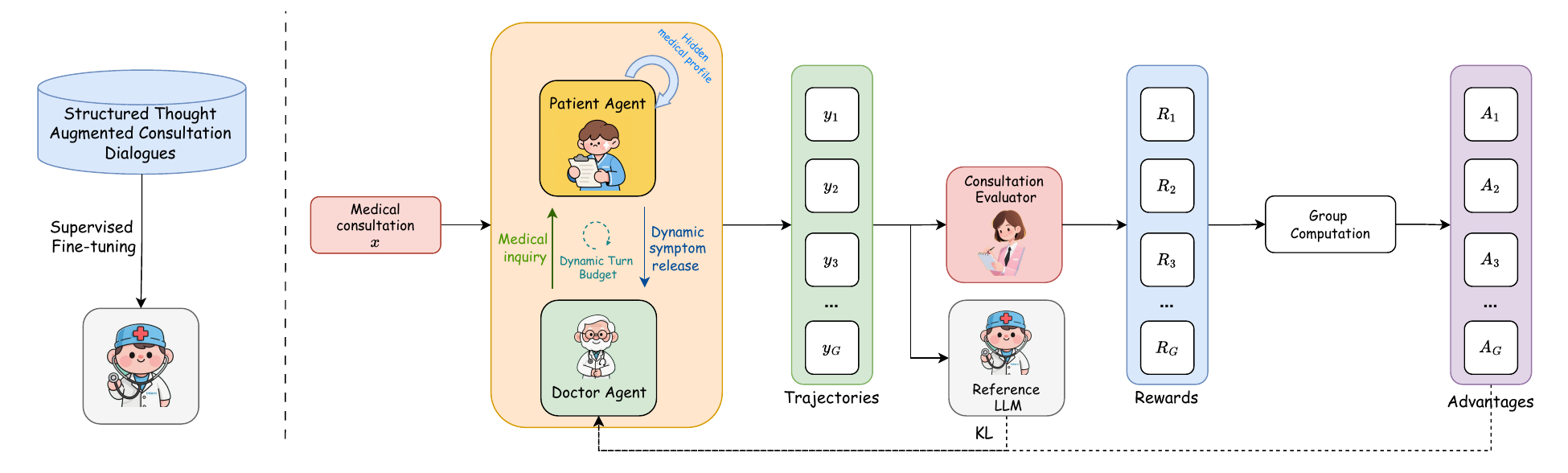}
        \caption{}
        \label{fig:framework}
    \end{subfigure}
    
    % \vspace{1em}
    
    % ============ 第2行: 图(b)和图(c)并列 ============
    \begin{minipage}[t]{0.34\textwidth}
        \vspace{0pt}
        % 图(b): 疾病分布统计
        \begin{subfigure}[t]{\textwidth}
            \centering
            \includegraphics[width=\textwidth]{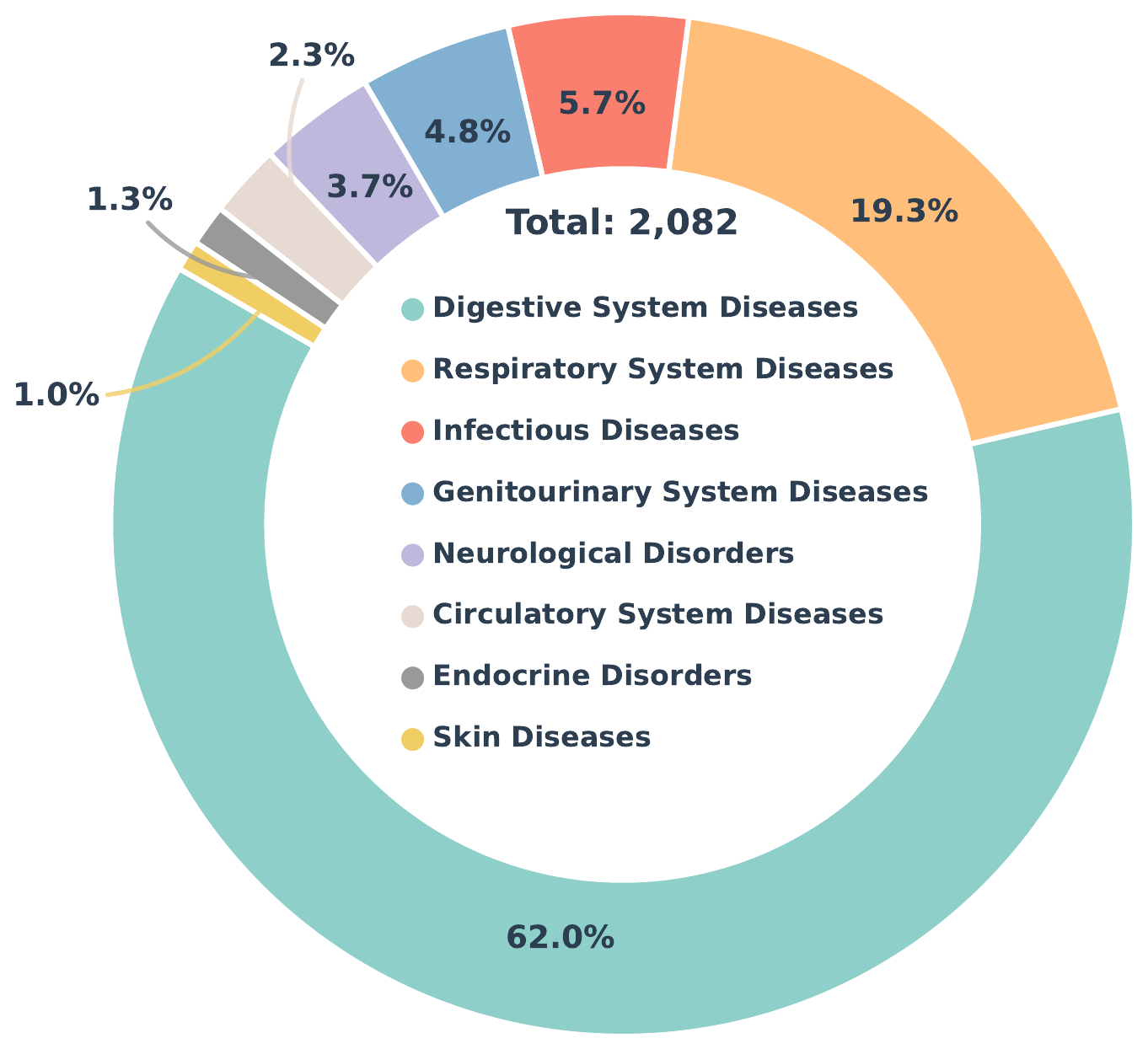}
            \caption{}
            \label{fig:disease_dist}
        \end{subfigure}
    \end{minipage}%
    \hfill
\begin{minipage}[t]{0.64\textwidth}
    \vspace{0pt}
    \vspace{0.5em}
    % 图(c): 数据集对比表格
    \begin{subfigure}[t]{\textwidth}
        \centering
        \resizebox{\columnwidth}{!}{%
            \begin{tabular}{@{}lccccc@{}}
                \toprule
                \textbf{Dataset} & \makecell{\textbf{Dynamic}\\\textbf{Interaction}} & \makecell{\textbf{Real-world}\\\textbf{Scenario}} & \makecell{\textbf{Patient}\\\textbf{Profile}} & \makecell{\textbf{Multi-}\\\textbf{turn}}  \\
                \midrule
                PubMedQA \cite{jin2019pubmedqa} & \xmark & \cmark & \xmark & \xmark  \\
                BioASQ-QA \cite{krithara2023bioasq} & \xmark & \cmark & \xmark & \xmark  \\
                HealthBench \cite{arora2025healthbench} & \cmark & \xmark & \xmark & \cmark  \\
                MedDialog \cite{he2020meddialog} & \xmark & \cmark & \xmark & \cmark \\
                IMCS21 \cite{10.1093/bioinformatics/btac817} & \xmark & \cmark & \cmark & \xmark  \\
                MedDG \cite{liu2022meddg} & \xmark & \cmark & \cmark & \xmark  \\
                CHIP-MDCFNP \cite{zhang2021cblue} & \xmark & \cmark & \cmark & \xmark  \\
                CRAFT-MD \cite{johri2025craftmd} & \cmark & \xmark & \cmark & \cmark  \\
                \midrule
                \textbf{MTMedDialog} & \cmark & \cmark & \cmark & \cmark \\
                
                \bottomrule
            \end{tabular}%
        }
        \vspace{0.1em}
        \caption{}
        \label{fig:dataset_comparison}
    \end{subfigure}
\end{minipage}
    
    % \vspace{1em}
    
    % ============ 第3行: 图(d) 占满整行 ============
    % \begin{subfigure}[t]{\textwidth}
    %     \centering
    %     \includegraphics[width=\textwidth]{recommendation_comparison.pdf}
    %     % 可以调整宽度: 0.6\textwidth, 0.7\textwidth, 0.8\textwidth
    %     \caption{}
    %     \label{fig:dataset_features}
    % \end{subfigure}
    %     \begin{subfigure}[b]{\columnwidth}
    %     \centering
    %     \includegraphics[width=\linewidth]{recommendation_comparison.pdf}
    %     \caption{}
    %     \label{fig:recommendation_comparison}
    % \end{subfigure}
    
    \caption{(a) The multi-agent collaborative reinforcement learning framework for \Ours{}. During the rollout stage, multi-turn interactions are conducted between the doctor agent and the patient agent. (b) Distribution of samples across eight major disease categories in test sets. (c) Feature comparison between MTMedDialog and existing medical dialogue datasets.}
    \label{fig:dataset_overview}
\end{figure*}

We propose DoctorAgent-RL, a multi-agent collaborative reinforcement learning framework that models medical consultation as a Markov Decision Process (MDP), as illustrated in Figure~\ref{fig:dataset_overview}(a). The system comprises three synergistic components that interact dynamically throughout the consultation process. The \textbf{Doctor Agent} serves as the primary decision-maker, learning strategic questioning policies to progressively elicit key patient information across multi-turn dialogues rather than memorizing diagnostic patterns. The \textbf{Patient Agent} simulates realistic symptom disclosure based on comprehensive hidden medical profiles, enabling dynamic information revelation conditioned on the doctor's queries instead of following predetermined scripts. The \textbf{Consultation Evaluator} acts as a neutral arbiter, providing multi-dimensional rewards that assess diagnostic accuracy, information acquisition efficiency, and protocol compliance, thereby guiding the doctor agent's policy optimization.

Through this framework, we train a doctor agent model, referred to as DoctorAgent-RL, based on Qwen2.5-7B-Instruct. Training follows a two-stage paradigm. First, supervised fine-tuning (SFT) on 1,000 reasoning-augmented dialogues establishes behavioral baselines through imitation learning, equipping the agent with fundamental consultation knowledge. Second, reinforcement learning via GRPO refines the doctor agent's questioning strategies through interactive feedback from both the patient agent and the consultation evaluator. To enhance robustness and adaptability, we introduce a dynamic turn budget mechanism during RL training, where each episode is assigned a random dialogue length constraint, encouraging the doctor agent to efficiently gather critical information under varying time pressures that mirror real clinical scenarios.
% \begin{figure*}[t]
%     \centering
%     \includegraphics[width=1.0\textwidth]{framework.pdf} % 图片路径
%     \caption{The multi-agent collaborative reinforcement learning framework for \Ours{}. During the rollout stage, multi-turn interactions are conducted between the doctor agent and the patient agent.} % 标题
%     \label{fig:framework} % 标签（用于交叉引用）
% \end{figure*}

%可以加入我们数据集的统计以及与其他数据集的对比，以及我们数据集的具体特点

\subsection{MTMedDialog: A Dynamic Multi-Turn Consultation Dataset}
\label{sec:dataset}
To support high-fidelity simulation of clinical interactions, we constructed MTMedDialog, comprising 8,086 training and 2,082 test samples across eight major disease categories: Digestive System Diseases, Respiratory System Diseases, Infectious Diseases, Genitourinary System Diseases, Neurological Disorders, Circulatory System Diseases, Endocrine Disorders, and Skin Diseases, as shown in Figure~\ref{fig:dataset_overview}(b) for test set distribution. A key distinction from existing datasets, as illustrated in Figure~\ref{fig:dataset_overview}(c), is that MTMedDialog consists of multi-turn dialogues derived from real-world clinical scenarios through data augmentation, where an LLM-based patient agent generates dynamic, contextually appropriate responses rather than following static dialogue transcripts or predetermined scripts. Each dialogue is grounded in a comprehensive hidden patient profile, enabling the patient agent to progressively reveal symptom information turn-by-turn in response to the doctor agent's strategic questions. This design faithfully mimics the uncertainty and interactive nature of real clinical consultations, providing an essential prerequisite for training agents that learn adaptive questioning strategies through multi-turn interactive simulation.

\subsection{Comparative Performance Evaluation of Multi-Model Approaches on MTMedDialog}

\begin{figure*}
    \centering
    \begin{subfigure}[b]{0.49\columnwidth}
        \includegraphics[width=\linewidth]{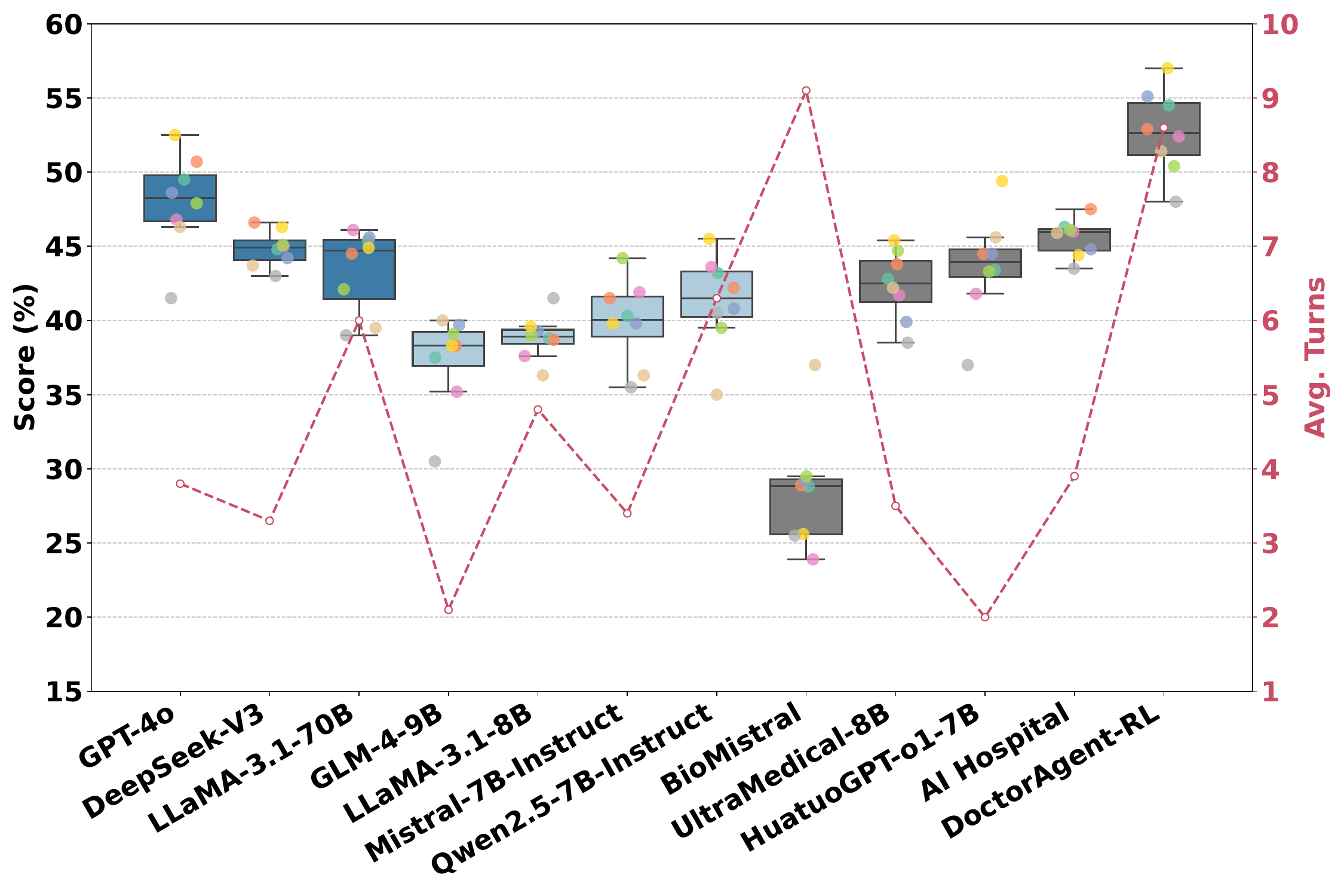}
        \caption{}
        \label{fig:performance_box_plot_2}
    \end{subfigure}
    \hfill
    \begin{subfigure}[b]{0.49\columnwidth}
        \includegraphics[width=\linewidth]{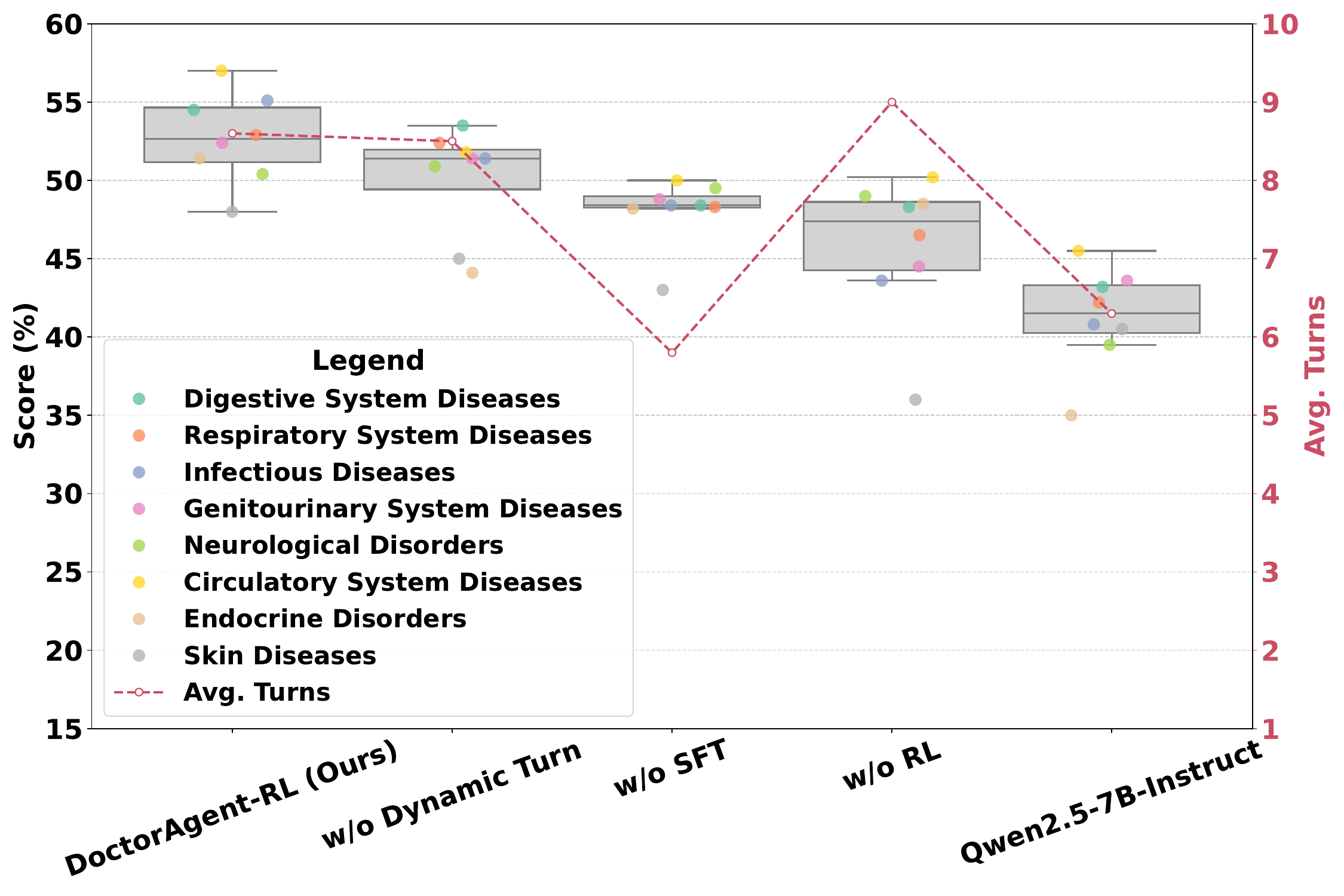}
        \caption{}
        \label{fig:ablation_study_box_and_scatter_plot_2}
    \end{subfigure}
    
    % \vspace{0.5em}
    \begin{subfigure}[b]{\columnwidth}
        \centering
        \includegraphics[width=\linewidth]{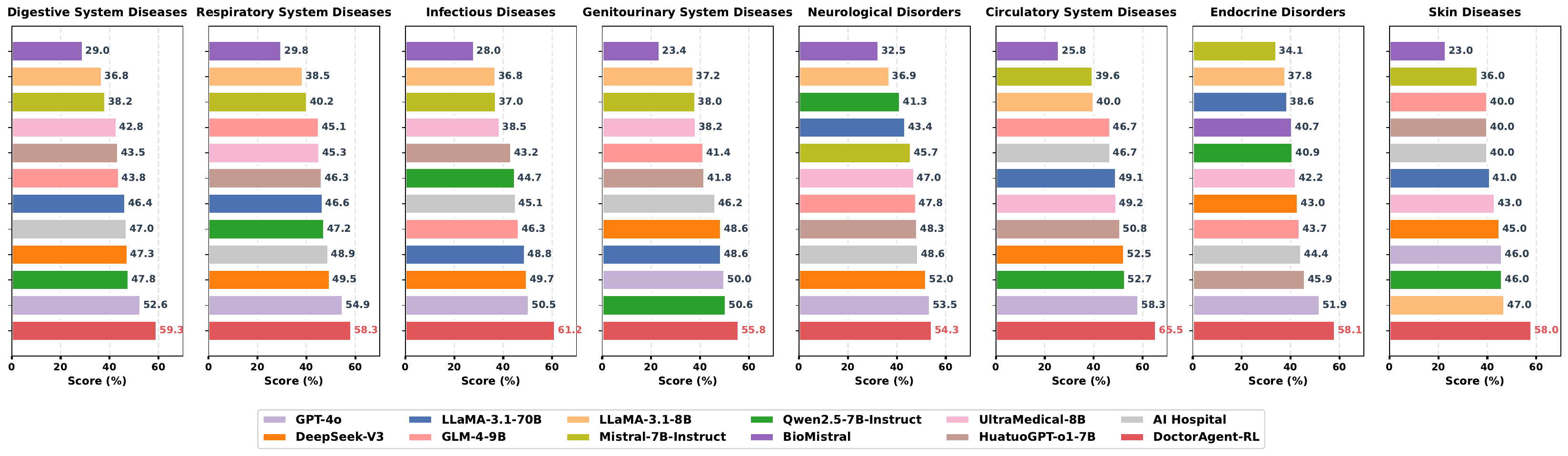}
        \caption{}
        \label{fig:diagnostic_comparison}
    \end{subfigure}
    
    % \vspace{0.5em}
    % \begin{subfigure}[b]{\columnwidth}
    %     \centering
    %     \includegraphics[width=\linewidth]{recommendation_comparison.pdf}
    %     \caption{}
    %     \label{fig:recommendation_comparison}
    % \end{subfigure}
    
    \vspace{0.5em}
    \begin{subfigure}[b]{0.49\columnwidth}
        \includegraphics[width=\linewidth]{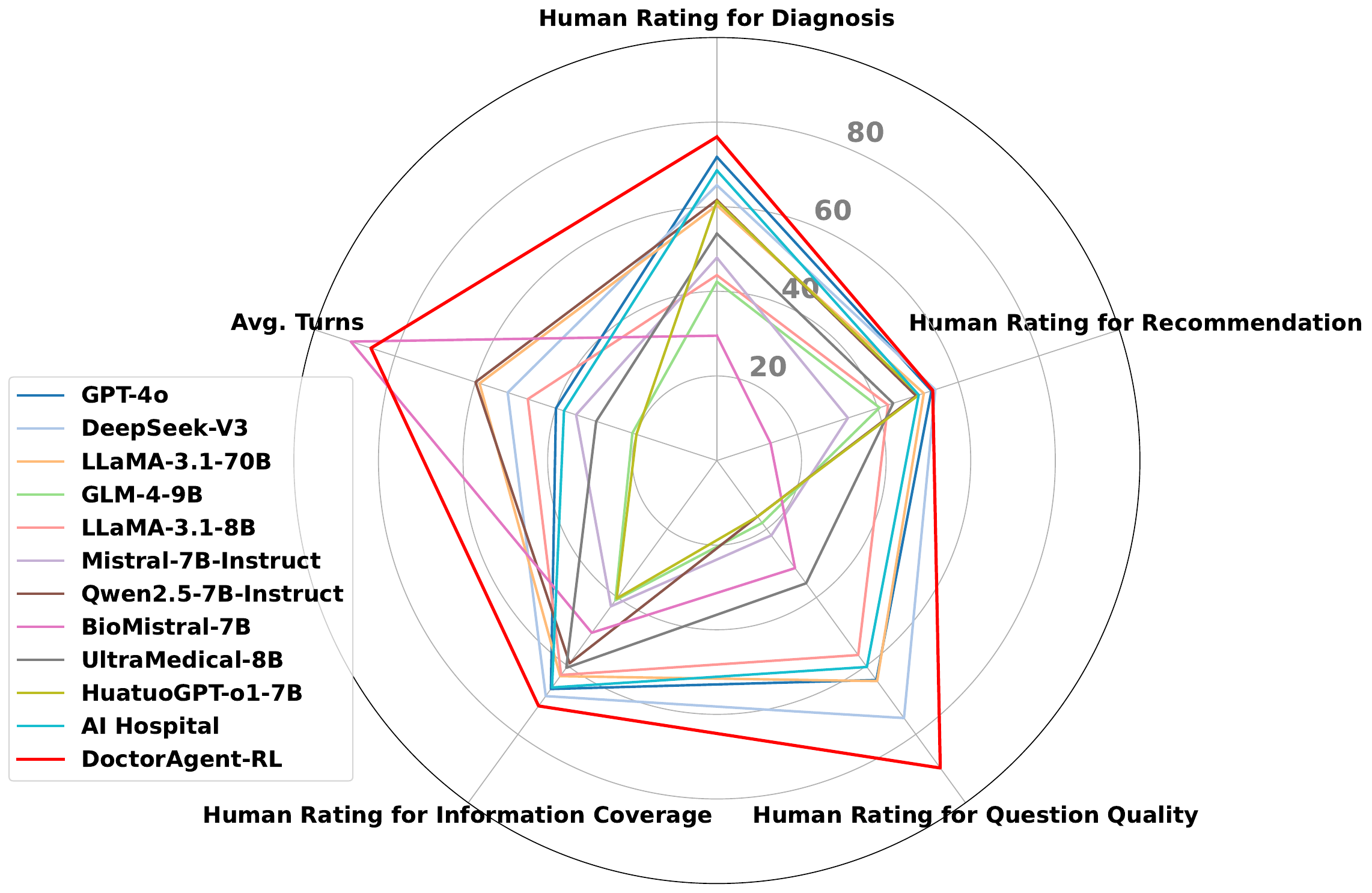}
        \caption{}
        \label{fig:MTMedDialog_performance}
    \end{subfigure}
    \hfill
    \begin{subfigure}[b]{0.49\columnwidth}
        \includegraphics[width=\linewidth]{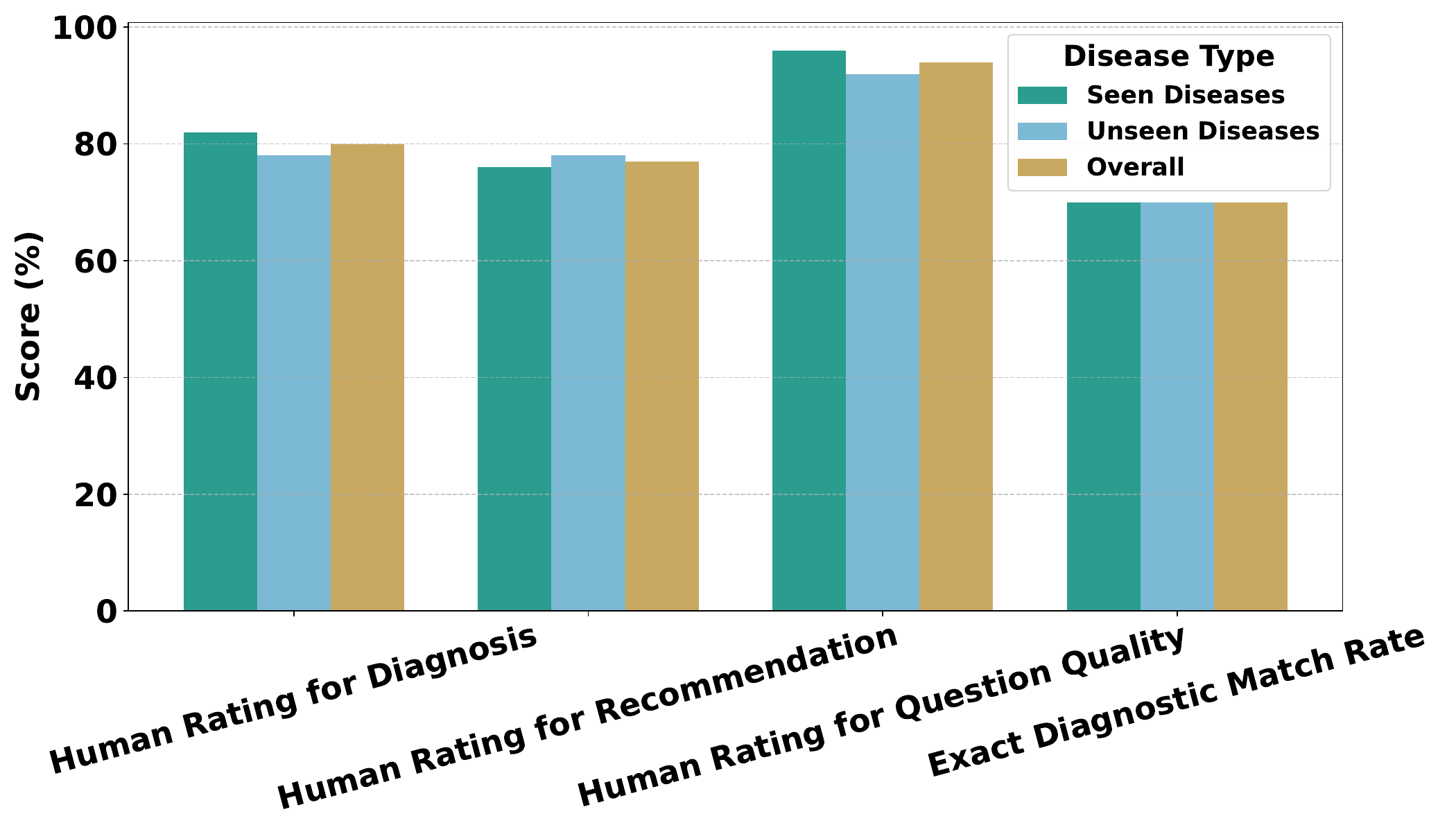}
        \caption{}
        \label{fig:doctoragent_performance_bar_chart}
    \end{subfigure}
    
    \caption{(a) Main results by disease category on MTMedDialog dataset, showing the average of diagnostic accuracy and recommendation accuracy scores. Dark blue boxes represent Frontier Models, light blue boxes represent Open-Source Base Models, and dark gray boxes represent Domain-Specific Models. (b) Performance comparison of different fine-tuning methods for Qwen2.5-7B-Instruct on MTMedDialog. (c) Detailed diagnostic performance comparison across eight disease categories on MTMedDialog. (d) Human randomly selected 100 data samples to compare the performance of different models on the MTMedDialog dataset. (e) Performance of DoctorAgent-RL in real-world interactive scenarios with actual patients.}
    \label{fig:combined_all}
\end{figure*}

The experimental results are shown in Figure~\ref{fig:combined_all}(a). The left Y-axis, Score (\%), represents Diagnosis and Recommendation Accuracy, a metric that uses a large language model (Qwen2.5-32B-Instruct) to assess the semantic consistency (on a 0-5 scale) between the model's output and the gold standard, which is then converted to a 100-point scale; the right Y-axis, Avg. Turns, represents the average interaction turns required for the model to reach a final diagnosis. As shown in the figure, \Ours{} achieved a comprehensive average score of 53.9\%, significantly outperforming all frontier, open-source base, and domain-specific models. More importantly, its box plot not only shows the highest median score, but its interquartile range is also positioned entirely above all competitors, indicating that our model performs more stably and with less variance across different disease types. As illustrated in Figure~\ref{fig:combined_all}(c), \Ours{} achieves the best performance across all eight disease categories in diagnostic accuracy, with no significant performance degradation in any specific domain. Unlike competing models that exhibit pronounced fluctuations between different disease types, our model demonstrates balanced performance across these diverse medical domains. This stable advantage on diseases that require in-depth consultation demonstrates its superior ability to simulate the clinical consultation process of doctors.

The figure clearly illustrates the relationship between performance and interaction frequency. For instance, the domain-specific model BioMistral has the highest average number of turns, yet its score box plot is at the lowest level and shows the widest distribution, indicating that its numerous questions are repetitive and ineffective. In contrast, some models (e.g., GLM-4, HuatuoGPT-o1-7B), despite having decent scores or very few turns, failed to acquire sufficient information through effective questioning or improperly combined multiple questions, which hindered the diagnosis of complex conditions. While frontier models strictly adhere to instructions, their lack of professional knowledge leads to questions that fail to address critical points. These findings provide strong evidence that effective diagnosis stems from the quality of questions, not the mere quantity of interactions.

The success of \Ours{} is attributed to its phased training strategy. The initial fine-tuning equips the model with systematic consultation knowledge, while the subsequent reinforcement learning allows it to move beyond imitating static patterns to truly learn and master a questioning methodology aimed at achieving an optimal diagnosis. The core of this methodology is not simply to increase the number of questions, but to actively guide the progressive emergence of key patient information in multi-turn dialogues through strategic, layered questioning. The relatively high Avg. Turns of \Ours{} is not ineffective or repetitive like in other models, but a direct reflection of its execution of this rigorous questioning methodology. Each interaction turn serves the final diagnostic goal, and therefore, this seemingly high interaction cost ultimately translates into the highest performance score among all models. This strongly proves that \Ours{} has learned a truly intelligent and efficient interaction strategy, and its interaction frequency is entirely valuable and reasonable.

\subsection{Component Contributions to Diagnostic Performance}
Experimental results demonstrate the significant advantages of our proposed two-stage optimization framework (SFT + RL) in medical dialogue tasks. As shown in Figure~\ref{fig:combined_all}(b), \Ours{} achieves a superior average score of 53.9, outperforming all baseline models. The key advantage stems from our phased strategy: (1) SFT on doctor-patient dialogues establishes reasonable questioning baselines through behavioral cloning, followed by (2) RL optimization that enables dynamic adjustment of questioning strategies for high-value information acquisition. 
Ablation studies on three critical components further validate our design:

\textbf{\textit{w/o} Dynamic\_Turn}: When trained with fixed budget of turns during RL, the model shows only a 1.2\% performance drop in matched scenarios but reveals strategy rigidity during inference—it mechanically adheres to the training budget of turns regardless of specified variations, proving impractical for real-world deployment.

\textbf{\textit{w/o} SFT}: Direct RL training without SFT initialization causes a 5.5\% average score degradation with the lowest average turns. While capable of planning effective questions for information gathering, the model demonstrates insufficient initiative in question generation due to the absence of behavioral cloning foundations.

\textbf{\textit{w/o} RL}: SFT-only training results in a 6.5\% lower average score despite having the highest turn count. The model memorizes question sequences from training data without truly understanding diagnostic logic, leading to mechanical questioning rather than strategic information acquisition.

The comprehensive performance of \Ours{} confirms each component's necessity: SFT establishes reliable behavioral baselines, RL injects dynamic decision-making capability, and adaptive turn mechanisms ensure strategic flexibility—together forming a reproducible paradigm for task-oriented medical dialogue optimization.

\subsection{Human Evaluation and In-depth Analysis of Multi-Model Approaches on MTMedDialog}
To conduct a more detailed evaluation of the comprehensive performance of different models in simulating real-world clinical consultation scenarios, we performed a human evaluation. Specifically, we randomly selected 100 samples from the MTMedDialog test set and then anonymized and shuffled the dialogues generated by all models for these samples. Evaluators, blinded to the model origins, independently scored the dialogues against the ground truth references. The scores were then mapped back to the respective models using their IDs to ensure the objectivity and fairness of the evaluation. Evaluation dimensions included Diagnostic Accuracy, Recommendation Reasonableness, Question Quality, and Information Coverage. The detailed results are presented in Figure \ref{fig:combined_all} (d).

The human evaluation results clearly indicate that our proposed DoctorAgent-RL outperforms all other models, achieving the highest scores in the three core metrics of Diagnostic Accuracy, Question Quality, and Information Coverage. It comprehensively surpassed frontier closed-source models, open-source base models, and other domain-specific models. This demonstrates that our method not only excels in the final diagnostic task but also holds a significant advantage in the strategic effectiveness of the consultation process.

In contrast, other models generally showed significant deficiencies in instruction following and interaction quality. Frontier models, representing the state-of-the-art in large language models, demonstrated strong overall capabilities. Among them, GPT-4o \cite{hurst2024gpt} achieved high scores across the board, making it the best alternative to our model. DeepSeek-V3 \cite{guo2025deepseek} excelled in question quality but sometimes violated the one-question-per-turn instruction. LLaMA-3.1-70B \cite{grattafiori2024llama} showed stable performance but lagged slightly behind the other two, similarly posing multiple questions in a single turn. While this increased information coverage, it also highlighted its weakness in instruction following. Open-source base models commonly suffered from poor instruction adherence and limited medical knowledge. For instance, Qwen2.5-7B-Instruct tended to provide a diagnosis directly rather than gathering information through questions and repeatedly violated the single-question rule. Mistral-7B-Instruct \cite{mistral7b} had the worst performance in instruction following, often failing to ask questions or producing repetitive, irrelevant output. Although LLaMA-3.1-8B \cite{grattafiori2024llama} could formulate relatively professional questions, its diagnostic accuracy was low due to insufficient internal knowledge. The advantages of DoctorAgent-RL were particularly prominent among domain-specific models. Other models in this category exhibited numerous issues. HuatuoGPT-o1-7B \cite{chen2024huatuogpt} and BioMistral-7B \cite{labrak2024biomistral} displayed severe problems with instruction following, even exhibiting "role reversal" by mimicking patient symptom descriptions. HuatuoGPT-o1-7B rarely asked questions and often included its overthinking process in the answers. BioMistral-7B had the highest average number of turns, but this was due to incessant, repetitive, and irrelevant questioning, leading to inefficiency and the lowest scores across all metrics. This phenomenon strongly proves that the key to effective multi-turn dialogue is question quality, not quantity. UltraMedical-8B \cite{zhang2024ultramedical} severely violated the single-question instruction by asking five to six questions at once, resulting in too few turns for in-depth, progressive information gathering. AI Hospital, a relatively high-performing domain-specific model, achieved performance close to that of frontier models but still showed a clear gap compared to our model.

In summary, our DoctorAgent-RL model, through SFT and RL, has learned to conduct high-quality, multi-turn interactions while adhering to the one question at a time instruction. The model is able to act like a real doctor, effectively gathering key information through a progressive and logical chain of questions, thereby achieving the highest diagnostic accuracy and information coverage.

\subsection{Performance Validation in Real Clinical Scenarios}
Automated evaluations and testing based on LLM-simulated patient datasets struggle to fully capture a model’s actual performance in real, dynamic conversational interactions. To comprehensively assess the end-to-end capabilities of DoctorAgent-RL in real clinical scenarios, we designed and conducted a real-world human-machine interaction evaluation. For this purpose, we recruited 20 real patients, ranging in age from their 20s to 80s, to engage in real-time online consultations with our model regarding their actual health conditions, which covered 15 common disease types. Subsequently, professional evaluators scored the complete dialogue process across multiple dimensions.

As detailed in Figure \ref{fig:combined_all} (e), DoctorAgent-RL demonstrated exceptional performance in these real-world interaction tests. The model achieved consistently high scores for both diagnostic accuracy and recommendation rationality, affirming its capacity to make reliable clinical judgments based on dynamically gathered information. Furthermore, the high scores in question quality, coupled with an efficient number of average dialogue turns, indicate that the model's path to achieving high accuracy is both clinically logical and effective.

To further scrutinize the model's clinical precision, we calculated the Exact Diagnostic Match Rate (ED), which measures the percentage of diagnoses that perfectly matched the ground truth. DoctorAgent-RL achieved a rate of 70\%, signifying that in the vast majority of interactions, it not only provided a reasonable assessment but correctly identified the patient's specific condition. This high rate of exact matches, combined with the strong overall performance scores, confirms that DoctorAgent-RL possesses the ability to stably, efficiently, and accurately complete clinical dialogue tasks in real-world interactive scenarios.

% \subsection{Ethics statement}
% This study involving human participants was reviewed and approved by the [Name of your Institution's Ethics Committee or IRB]. All 20 patients provided their written informed consent to participate in this study. The consent form explicitly stated that the conversational agent is an AI model for research purposes only, not a certified medical professional, and that any information provided should not be considered medical advice. All collected dialogue data was rigorously anonymized to protect participant privacy.

% \subsection{Ethical Considerations}
% This study involved interactions with 20 real patients. Prior to any interaction, we fully informed each participant about the study's purpose, procedures, and data usage, and obtained their written informed consent. The consent form emphasized that the conversational agent was an AI model for research purposes only, not a certified medical professional, and that any information provided should not be considered medical advice. All data collected from these interactions was rigorously anonymized to protect participant privacy and confidentiality. All data was used solely for the analysis in this test and was not used in any model training processes.

\subsection{In-depth Analysis of Generalization and General-Purpose Capabilities}

Having validated the model's overall performance in real-world scenarios, we proceeded with an in-depth analysis focusing on two dimensions that are critical for evaluating the maturity of a specialized LLM: its generalization ability and its general-purpose capabilities. The former assesses the model's capacity to handle unseen clinical cases, while the latter evaluates whether its foundational conversational skills remain intact after domain-specific training.

We began by analyzing the performance discrepancies between disease types the DoctorAgent-RL had seen versus those it had not been exposed to during its training phase. As illustrated in Figure \ref{fig:combined_all}(e), the results for both groups of cases showed negligible differences across core metrics. This suggests that the model has acquired a transferable clinical reasoning framework rather than relying on rote memorization of disease-specific patterns, thereby demonstrating strong generalization capabilities.

\begin{figure*}
    \centering
    % 新的图a
    \begin{subfigure}[b]{\columnwidth}
        \centering
        \includegraphics[width=0.8\linewidth]{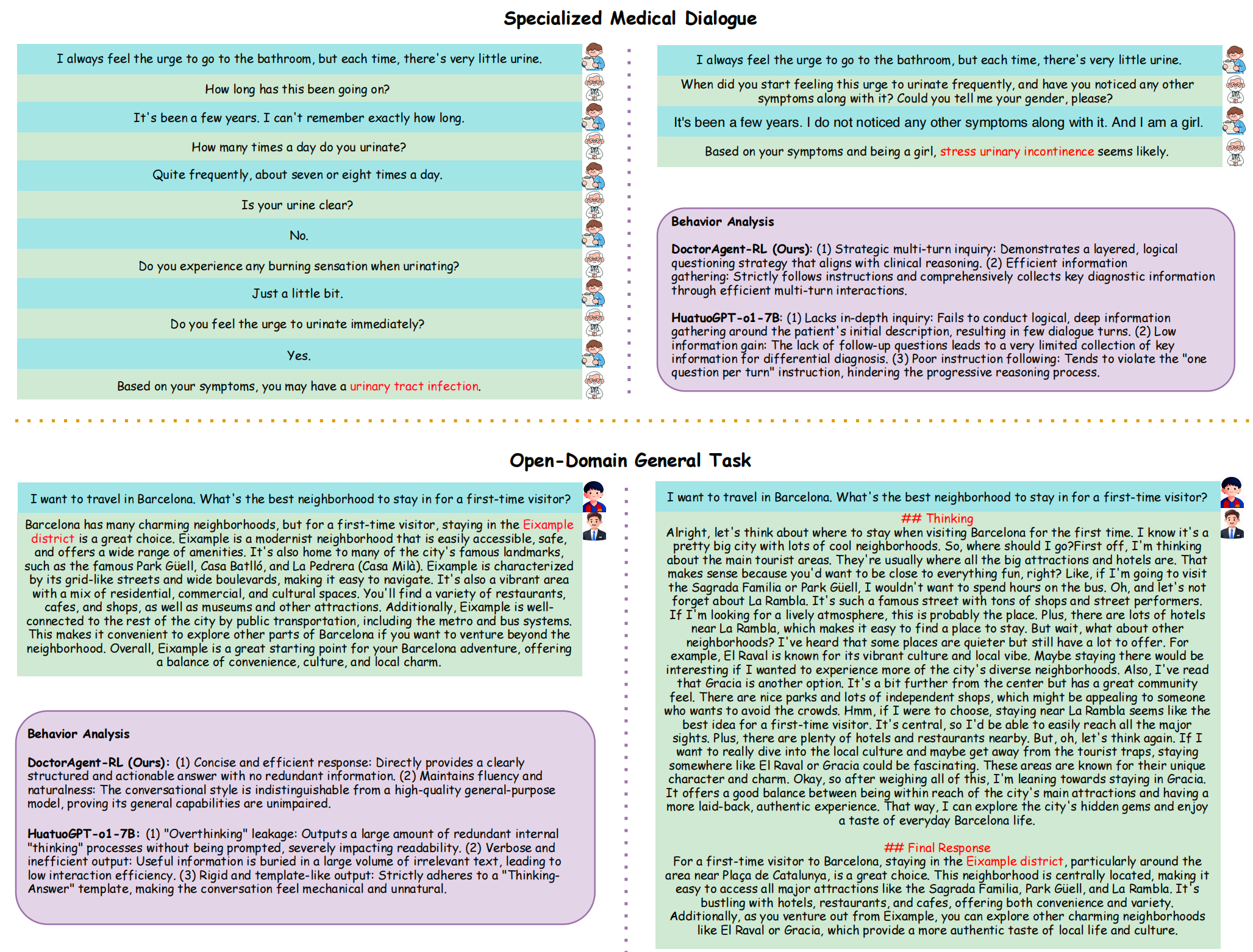} % 请替换为你的新图文件名
        \caption{}
        \label{fig:new_a}
    \end{subfigure}

    \vspace{-0.5em} % 可选的垂直间距，让三图并排与上图a分开

    % 原来的图a（现在变成图b）和bc三图并排
    \begin{subfigure}[b]{0.34\linewidth} % 原来的图a
        \centering
        \includegraphics[width=\linewidth]{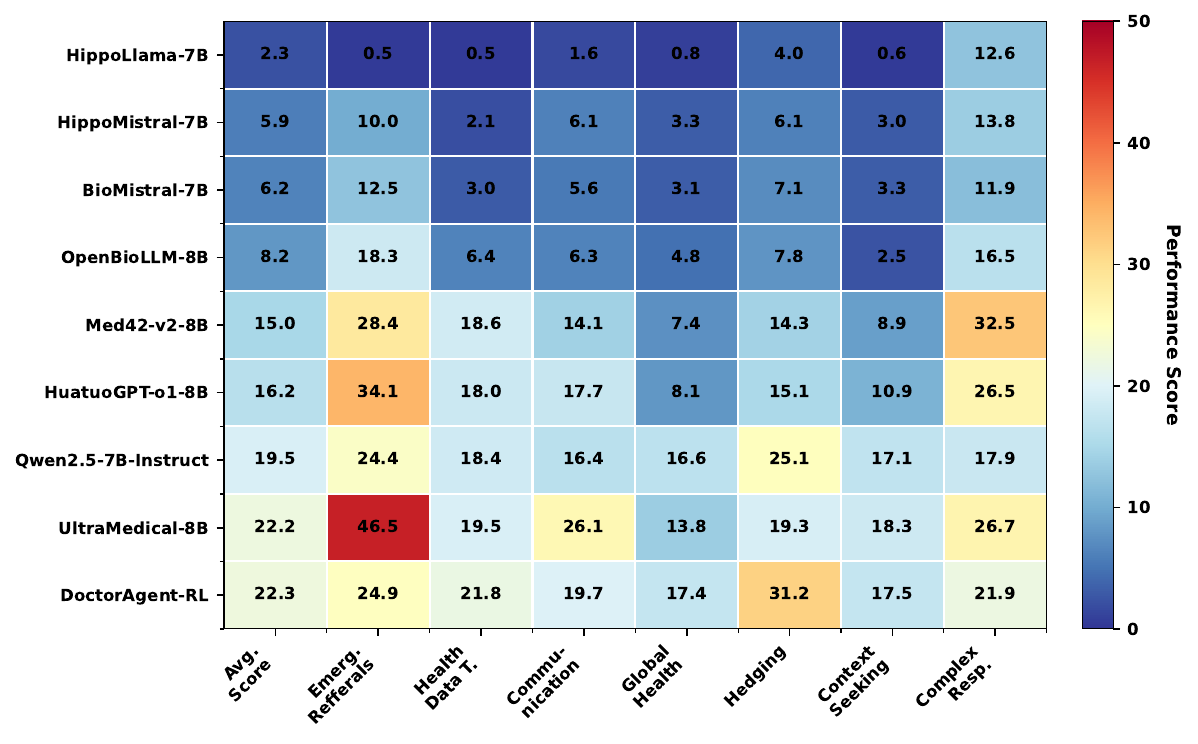}
        \caption{}
        \label{fig:heatmap_updated}
    \end{subfigure}
    \hfill
    % \begin{subfigure}[b]{0.30\linewidth} % 原来的图b
    %     \includegraphics[width=\linewidth]{performance_vs_budget.pdf}
    %     \caption{}
    %     \label{fig:performance_vs_budget}
    % \end{subfigure}
    \begin{subfigure}[b]{0.30\linewidth}
    \raisebox{0.2cm}{% 调整这个数值，正数上移，负数下移
        \includegraphics[width=\linewidth]{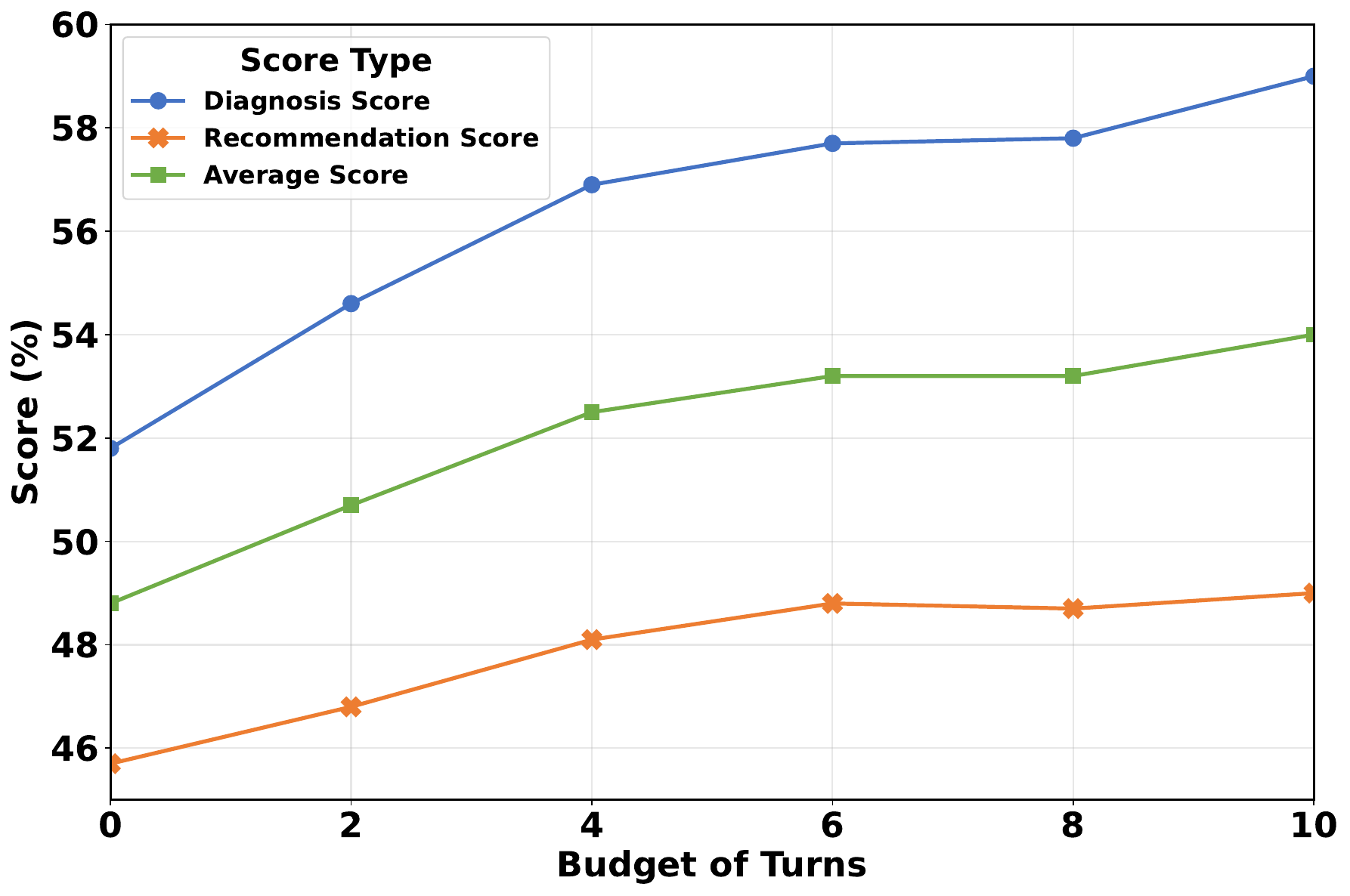}
    }
    \caption{}
    \label{fig:patient_agent_performance_bar_chart}
\end{subfigure}
    \hfill
    \begin{subfigure}[b]{0.32\linewidth} % 原来的图c

        \includegraphics[width=\linewidth]{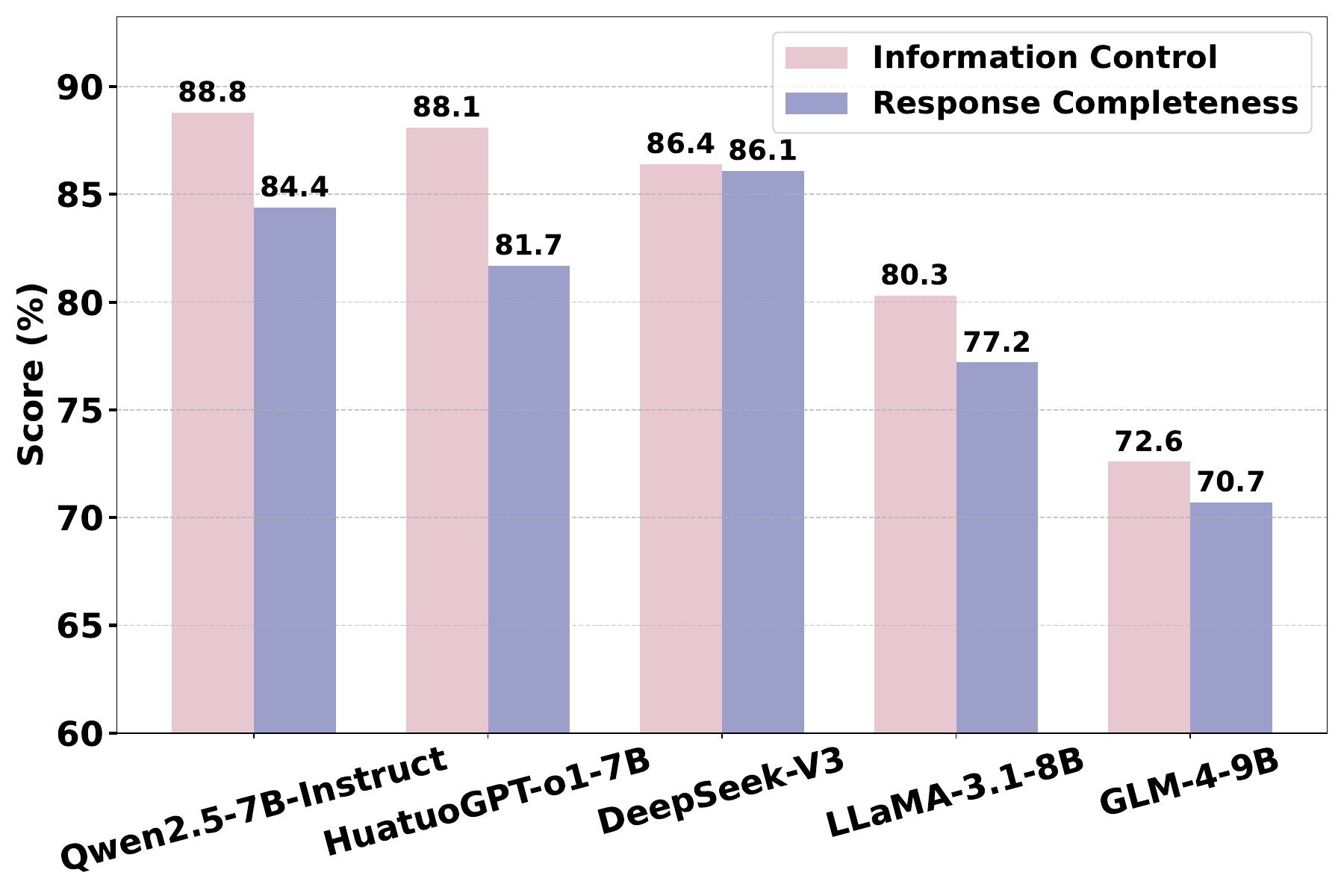}
        \caption{}
        \label{fig:patient_agent_performance_bar_chart}
    \end{subfigure}

    \caption{(a) Qualitative comparison of dialogue outputs from DoctorAgent-RL (left column) and HuatuoGPT-o1-7B (right column) during interactions with a human user. (b) Performance comparison on HealthBench benchmark across eight evaluation dimensions. Partial experimental data sourced from \cite{lai2025doctor}. (c) Comparative analysis of Diagnosis Score, Recommendation Score, and the resulting Average Score of \Ours{} on MTMedDialog at different levels of the Budget of Turns. (d) Performance comparison of different models in simulating patient agent on MTMedDialog.}
    \label{fig:patient_agent_performance}
\end{figure*}

To further evaluate the robustness and general-purpose capabilities of DoctorAgent-RL, we conducted a qualitative analysis where human evaluators interacted with multiple models across a range of scenarios, covering specialized medical inquiries, open-domain tasks (e.g., travel planning), and role-playing exercises. In our tests, we observed significant behavioral differences across different training paradigms. For instance, domain-specific SFT LLMs like BioMistral exhibited severe instruction-following deficiencies, leading to unsatisfactory performance in both medical and general tasks; its effectiveness is likely confined to a narrow distribution highly similar to its training data. In contrast, a domain-specific RL LLM like HuatuoGPT-o1-7B, despite performing well on the medical task, showed clear signs of being over-tuned, resulting in rigid, template-like outputs and a severe degradation of its general-purpose conversational abilities.
In contrast, DoctorAgent-RL successfully avoids this issue. It not only excels in specialized medical dialogues but also fully preserves the powerful general-purpose conversational skills of its base model, remaining fluent and efficient in non-medical tasks. The model is also robust in terms of role consistency; when instructed to switch roles and act as a patient, it can describe symptoms naturally without any confusion from its core doctor training. Figure \ref{fig:patient_agent_performance}(a) shows a set of representative dialogues that vividly illustrates the behavioral differences between our model and HuatuoGPT-o1-7B on both a medical and a travel planning task. Based on this qualitative evaluation, we conclude that DoctorAgent-RL is an intelligent agent that achieves deep domain specialization while fully preserving its general-purpose capabilities.

\subsection{Generalization Assessment on HealthBench}
To further evaluate the generalization capability of DoctorAgent-RL on unseen medical data distributions, we conducted comprehensive evaluations on HealthBench \cite{arora2025healthbench}, a standardized benchmark for medical AI systems covering eight evaluation dimensions: Average Score, Emergency Referrals, Health Data Interpretation, Communication Skills, Global Health Knowledge, Hedging Appropriateness, Context-Seeking Behavior, and Complex Response Handling. Following the official HealthBench evaluation protocol, all models were assessed using GPT-4.1 as the automated scoring mechanism to ensure methodological consistency. Notably, both the Qwen2.5-7B-Instruct and our DoctorAgent-RL were evaluated under identical conditions using the official HealthBench testing infrastructure.

The experimental results are presented in Figure~\ref{fig:patient_agent_performance}(b). DoctorAgent-RL achieved an average score of 22.3\%, ranking first among all evaluated open-source small-scale models. This strong performance on unseen data distributions is particularly noteworthy given that DoctorAgent-RL was specifically trained on MTMedDialog, a multi-turn medical consultation dataset emphasizing active information gathering through strategic questioning and progressive diagnosis. The substantial improvement over the base model and competitive results compared to models trained on broader medical corpora demonstrate that our reinforcement learning framework enables the agent to learn transferable clinical reasoning patterns rather than merely memorizing consultation templates. The model's ability to maintain robust performance across diverse evaluation dimensions—ranging from emergency triage to nuanced communication skills—suggests that the strategic information-gathering methodology acquired during training generalizes effectively beyond the specific consultation scenarios encountered during development. These results provide compelling evidence that training medical AI systems to actively construct knowledge through dynamic interaction, rather than passively reproducing existing knowledge, leads to more generalizable and clinically versatile capabilities even when evaluated on data distributions substantially different from the training corpus.

\subsection{Impact Analysis of Budget of Turns on Diagnosis and Recommendation Performance}
The experimental results shown in Figure \ref{fig:patient_agent_performance} (c) demonstrate that as the budget of turns increases, the average performance of diagnosis and recommendation exhibits a distinct two-phase characteristic: in the initial phase (low-turn range), performance rises rapidly with additional turns, primarily due to the multi-turn dialogue mechanism enabling the LLM to progressively collect and refine patient information through iterative questioning; in the intermediate phase (medium-to-high turn range), the performance curve's slope noticeably flattens as the valuable information patients can provide gradually becomes saturated, making it difficult for the LLM to extract more meaningful information through additional questioning. Nevertheless, on the whole, a larger budget of turns still leads to better performance, as additional interaction opportunities can capture potential subtle information differences. It is worth noting that diagnostic performance consistently outperforms recommendation performance, as diagnosis tasks can progressively confirm symptom characteristics through multi-turn interactions, while recommendation tasks rely more on established medical knowledge bases, leaving relatively limited room for performance improvement.

\subsection{Patient Agent Behavior Analysis on MTMedDialog}

Figure~\ref{fig:patient_agent_performance} (d) presents the evaluation results of the effectiveness of the patient agent in simulating real-world medical interactions. The assessment employs a scoring system based on DeepSeek-V3, evaluating two key metrics: (1) Information Control, which quantifies unsolicited information disclosure with a baseline score of 1.0 (indicating perfect compliance) and deducts 0.2 points per unauthorized disclosure instance; and (2) Response Completeness, which evaluates critical information omission against doctor queries using an initial score of 1.0, penalized by 0.2 points per essential detail omission. Both metrics are ultimately converted to a percentage-based rating.
In the dimension of information control, Qwen2.5-7B demonstrates optimal compliance performance, precisely constraining output content and effectively avoiding the leakage of irrelevant information. In terms of response completeness, while DeepSeek-V3 performs best, Qwen2.5-7B still maintains near-optimal levels of key information density required for doctor-patient dialogue. Considering both model performance and cost-effectiveness in training, we ultimately selected Qwen2.5-7B as the implementation solution for the patient agent. For a more detailed description of the dialogue flow and rating prompts, please refer to Supplementary Material.

\section{Discussion}\label{sec12}
This study successfully designed and validated a multi-agent reinforcement learning framework named DoctorAgent-RL, aimed at addressing the core challenges faced by current large language models in real-world, multi-turn clinical consultations. By modeling clinical diagnosis as a dynamic decision-making process and training the agent to master a ``questioning methodology" for optimal diagnosis, our model has demonstrated exceptional performance across multiple evaluations.

Our most significant findings come from a multi-faceted and rigorous evaluation system. In both automated evaluations based on our custom-built MTMedDialog dataset and in blinded human assessments, DoctorAgent-RL significantly outperformed all baselines, including frontier closed-source and domain-specific models, on key metrics such as diagnostic accuracy, recommendation rationality, and question quality. Critically, in real-time interaction tests with 20 real patients, our agent achieved a 70\% exact diagnostic match rate. This result provides powerful evidence that our framework is not only effective in simulated environments but, more importantly, possesses the robust practical capability to stably and efficiently complete clinical dialogue tasks in unpredictable real-world scenarios, successfully bridging the critical sim-to-real gap.

Compared to existing work, the advantages of DoctorAgent-RL are evident. Traditional supervised fine-tuning models, such as BioMistral \cite{labrak2024biomistral}, exhibit poor performance despite a high number of interaction turns due to their inability to comprehend diagnostic logic, confirming the limitations of superficial imitation \cite{liu2025interactive}. In contrast to other frontier multi-agent or reinforcement learning systems (e.g., AMIE, HuatuoGPT-01-7B) \cite{tu2025towards,chen2024huatuogpt}, our two-stage training paradigm (SFT+RL) and the MTMedDialog dataset, designed specifically for dynamic interaction, enable the model to learn a transferable clinical reasoning framework rather than merely "overfitting" to specific disease patterns. Our qualitative analysis also indicates that DoctorAgent-RL successfully preserves the general-purpose conversational abilities of its base model while acquiring deep domain expertise, thus avoiding the "catastrophic forgetting" common in many domain-specific models.

Despite these encouraging results, we also recognize the limitations of this study. First, although the MTMedDialog dataset is innovative in its support for dynamic interaction, it currently covers only eight major disease categories. This may limit the model's generalizability across a broader spectrum of diseases and different cultural contexts. Second, our real-patient validation study, while a crucial step in verifying the model's practical capabilities, has a relatively small sample size (20 patients). Larger and more diverse clinical trials will be necessary to further confirm its safety and efficacy. Furthermore, while our high-fidelity patient agent can effectively simulate dialogues, it cannot fully replicate the complex psychological states, emotional variations, and descriptive inconsistencies of real human patients. Lastly, the current framework does not yet integrate multimodal information (e.g., medical imaging), which is essential for real-world diagnosis.

Looking forward, our work opens up several research directions for developing next-generation clinical decision support systems. First, efforts can be dedicated to expanding and enriching the MTMedDialog dataset to include more disease types, languages, and cultural backgrounds. Second, conducting larger-scale, prospective clinical trials is a necessary step for deploying this model in real clinical settings. Future research should also explore the integration of multimodal data inputs into the model's decision-making loop. Additionally, enabling the model to explain its thought process to clinicians to enhance transparency and trustworthiness is another important topic for in-depth investigation.

In summary, DoctorAgent-RL not only technically surpasses existing medical dialogue systems but, more importantly, provides a clear and validated path for building a medical AI capable of dynamic, strategic information gathering. Our findings suggest that by positioning AI as a collaborative tool to augment, rather than replace, physicians' capabilities, we can hope to develop intelligent systems that can genuinely alleviate the pressure on global healthcare resources, reduce the risk of misdiagnosis, and ultimately improve the quality of patient care.

\section{Methods}\label{sec11}

\subsection{Task Formulation}
\label{sec:rl}

We model the multi-round medical consultation process as a multi-agent collaborative RL system, where the doctor agent serves as the main agent for strategy optimization, the patient agent acts as a collaborative counterpart to form a dynamic game relationship with it, and the consultation evaluator functions as a neutral arbiter that coordinates doctor-patient interactions and guides the optimization process of the doctor agent's strategy through well-designed reward mechanisms.

\subsubsection{Doctor Agent}
%需要检查策略函数，以及是否在这一节说明训练策略？
% The doctor agent, as the decision-making subject, maintains a state space $s_t \in \mathcal{S}$ containing the dialogue history $H_t \in \mathbb{R}^d$ (where $d$ is the embedding dimension), forming a complete representation of the consultation process. The action space $\mathcal{A} = \{a_{query}, a_{diagnose}\} \subset \mathbb{R}^m$ comprises two behavioral modes: generating medical inquiries and executing diagnostic decisions. 

As the decision-making doctor agent, its state space $s_t \in \mathcal{S}$ encompasses the dialogue history $H_t$, providing a comprehensive record of the consultation. The agent's actions are drawn from the action space $\mathcal{A} = \{a_{query}, a_{diagnose}\}$, which includes two distinct behaviors: generating medical inquiries and executing diagnostic decisions.

% The reward signal $r_t \in \mathbb{R}$ is provided by an independent consultation evaluator:
% \[
% r_t = \sum_{i=1}^3 w_i R_i(s_t,a_t), \quad \text{where } \sum w_i = 1
% \]
% with $R_1$ (diagnostic accuracy), $R_2$ (information
% acquisition rate), and $R_3$ (protocol compliance) as evaluation metrics.

Environmental dynamics are governed by the patient agent through state transitions:
\[
s_{t+1} \sim P(s_{t+1}|s_t,a_t)
\]

Here, $P$ represents the transition probability function, which quantifies the likelihood of moving to a new state $s_{t+1}$ given the current state $s_t$ and the action $a_t$ taken by the doctor agent.

The consultation continues until either a predefined limit on conversation turns is reached or the doctor agent provides a final diagnosis, at which point a complete dialogue history $H_T$ forms the consultation trajectory.

The reward signal $R \in \mathbb{R}$ for each trajectory is provided by an independent consultation evaluator (detailed in Section \ref{sec:reward}), considering diagnostic accuracy, information
acquisition efficiency, and protocol compliance as metrics.

% The policy $\pi_\theta: \mathcal{S} \rightarrow \mathcal{A}$, implemented via a large language model, is optimized using PPO:
% \[
% \theta_{new} = \arg\max_\theta \mathbb{E}_{s,a\sim\pi_{old}} \left[ \min\left( \frac{\pi_\theta(a|s)}{\pi_{old}(a|s)} A_t, \text{clip}\left(\frac{\pi_\theta(a|s)}{\pi_{old}(a|s)}, 1-\epsilon, 1+\epsilon\right) A_t \right) \right]
% \]

% The generalized advantage estimator (GAE) computes:
% \[
% A_t = \sum_{l=0}^{T-t} (\gamma\lambda)^l \delta_{t+l}, \quad \delta_t = r_t + \gamma V_\phi(s_{t+1}) - V_\phi(s_t)
% \]

% Subject to KL divergence constraint:
% \[
% D_{KL}(\pi_\theta||\pi_{old}) \leq \delta, \quad \delta = 0.01
% \]

% The optimization objective is:
% \[
% \max_\theta \mathbb{E}\left[\sum_{t=0}^T \gamma^t r_t\right], \quad \gamma \in (0,1)
% \]

To enhance the stability of policy optimization and eliminate the requirement for an additional value function approximation, we propose Group Relative Policy Optimization \cite{shao2024deepseekmath} as our policy gradient algorithm. Unlike Proximal Policy Optimization \cite{schulman2017proximal}, GRPO employs the average reward of multiple sampled outputs as a baseline instead of relying on a learned value function. Specifically, for each patient's consultation \(x\), GRPO samples a set of trajectories ${y_1, y_2, \dots, y_G}$ through the interaction between the doctor agent $\pi_D$ and the patient agent $\pi_\text{p}$. The doctor agent, as the policy model, is then optimized by maximizing the following objective function:

{\scriptsize
\begin{multline}\label{eq:grpo}
\mathcal{J}_{GRPO}(\theta) = 
\mathbb{E}_{
    \substack{
        x \sim \mathcal{D} \\ 
        \{ y_i \}_{i=1}^{G} \sim \pi_{\mathrm{D_{old}}}( \cdot| x; \pi_\mathrm{p})
    }
}
\Bigg[
\frac{1}{G} \sum_{i=1}^{G} \frac{1}{\sum_{t=1}^{|y_i|}  I(y_{i,t})} 
\sum_{\substack{t=1 \\ I(y_{i,t})=1}}^{|y_i|} \\
\min \Bigg( 
\frac{\pi_{\mathrm{D}}(y_{i,t} | x, y_{i,<t})}{\pi_{\mathrm{D_{old}}}(y_{i,t} | x, y_{i,<t})} \hat{A}_{i,t},\ 
\text{clip} \Bigg( \frac{\pi_{\mathrm{D}}(y_{i,t} | x, y_{i,<t})}{\pi_{\mathrm{D_{old}}}(y_{i,t} | x, y_{i,<t})}, 
1 - \epsilon, 1 + \epsilon \Bigg) \hat{A}_{i,t} 
\Bigg) 
- \beta \mathbb{D}_{\mathrm{KL}} \left[ \pi_{\mathrm{D}} || \pi_{\mathrm{D_{ref}}} \right]
\Bigg]
\end{multline}
}

% {\scriptsize
% \begin{align}\label{eq:grpo}
% \mathcal{J}_{GRPO}(\theta) = \, & 
% \mathbb{E}_{x \sim \mathcal{D}, \{ y_i \}_{i=1}^{G} \sim \pi_{\text{D}_\text{old}}( \cdot| x; \pi_\text{p})}
% \Bigg[
% \frac{1}{G} \sum_{i=1}^{G} \frac{1}{\sum_{t=1}^{|y_i|}  I(y_{i,t})} \sum_{t=1: I(y_{i,t})=1}^{|y_i|} 
% \min \Bigg( 
% \frac{\pi_{\text{D}}(y_{i,t} | x, y_{i,<t})}{\pi_{\text{D}_\text{old}}(y_{i,t} | x, y_{i,<t})} \hat{A}_{i,t}, 
% \nonumber \\[8pt] 
% & \hspace{120pt} \text{clip} \Bigg( \frac{\pi_{\text{D}}(y_{i,t} | x, y_{i,<t})}{\pi_{\text{D}_\text{old}}(y_{i,t} | x, y_{i,<t})}, 1 - \epsilon, 1 + \epsilon \Bigg) \hat{A}_{i,t} 
% \Bigg)
% - \beta \mathbb{D}_{KL} \left[ \pi_{\text{D}} || \pi_{\text{D}_{ref}} \right]
% \Bigg],
% \end{align}
% }
where $\epsilon$ and $\beta$ are hyperparameters, and $\hat{A}_{i,t}$ represents the advantage, computed based on the relative rewards of outputs within each group. Here, $I(y_{i,t}) = 1$ indicates that the token $y_{i,t}$ is generated by $\pi_{\text{D}}$. Since doctors cannot predict patients' symptoms in advance, during training, the responses of the patient agent are masked.

\subsubsection{Patient Agent}
The patient agent is implemented using Qwen2.5-7B-Instruct \cite{xu2025qwen2} and is incorporated into a two-phase dialogue simulation framework via carefully crafted prompt engineering. 

In the first phase, the system combines patient self-reports and multi-turn dialogue content to create case descriptions. Additionally, it augments potential symptom features using standard diagnostic results, thereby forming a more comprehensive hidden medical profile. This design effectively mitigates the symptom coverage issues stemming from incomplete doctor inquiries in traditional datasets. 

In the second phase, the patient agent utilizes dynamic symptom release strategies in response to the real-time queries of the doctor agent. It maintains strict pathological consistency while mimicking the natural variability in patients' symptom description granularity and the order of their complaints. The detailed prompt designs for both phases are presented in Supplementary Material. By retaining complete hidden case data, the patient agent ensures that its natural language responses adhere to clinical standards and are generated solely based on the dialogue history.

\subsubsection{Consultation Evaluator}
\label{sec:reward}
In our RL framework, guiding the doctor agent to master essential clinical diagnostic skills is paramount. To achieve this, we've designed a sophisticated Consultation Evaluator, acting as a multi-faceted reward system that assesses the agent's performance across critical dimensions of a medical consultation. This evaluator comprises three core components, each contributing to a comprehensive assessment of the agent's diagnostic capabilities and consultation conduct.

\textbf{Diagnostic Accuracy Reward.} The first pillar of our Consultation Evaluator focuses on the agent's diagnostic precision and treatment recommendations. To ensure the reliability of this evaluation and prevent any potential reward hacking, we employ a rule-based reward mechanism. This mechanism meticulously calculates the word-level F1 score between the doctor agent's predicted diagnosis and the gold-standard diagnosis, as well as for the recommended treatments. The formulation for this reward is as follows:

{\scriptsize
\begin{equation}
    R_{\text{accuracy}} = 5 \times \left( \text{F1}_{\text{diagnosis}} + \text{F1}_{\text{recommendation}} \right)
\end{equation}
}
The coefficient 5 serves to adjust the relative weight of this crucial reward signal within the overall evaluation. 
This design ensures a balanced and robust assessment of both the diagnostic output and the quality of suggested interventions, providing a stable foundation for learning.

\textbf{Information Acquisition Efficiency Reward.} 
A proficient doctor knows how to ask the right questions efficiently. To foster this skill in our doctor agent, the Consultation Evaluator incorporates a dynamic reward mechanism that promotes valuable questioning while discouraging repetitive or unhelpful queries. This reward is directly tied to the patient agent's feedback after each dialogue turn and accumulates throughout the interaction:

{\scriptsize
\begin{equation}
R_{\text{information}}^t = 
\begin{cases} 
1, & \text{if the patient agent answers normally} \\
-2, & \text{if the patient agent refuses to answer} 
\end{cases} 
\end{equation}
}

{\scriptsize
\begin{equation}
    R_{\text{information}} = \sum_{t} R_{\text{information}}^t
\end{equation}
}
Through this feedback loop, the model learns to optimize its questioning strategy, prioritizing the acquisition of diagnostically relevant information and refining its inquiry process.

\textbf{Protocol Compliance Reward.} Adherence to established clinical interview protocols is a hallmark of professional medical practice. To instill this discipline, our Consultation Evaluator includes a compliance reward. This component penalizes deviations from predefined norms and ensures that the agent completes the diagnostic process within specified limits:

{\scriptsize
\begin{equation}
R_{\text{compliance}}^t = 
\begin{cases} 
-2, & \text{Question Format Violation} \\
-5, & \text{No Diagnosis within Allowed Turns} \\
0, & \text{otherwise (i.e., protocol is followed)} 
\end{cases}
\end{equation}
}

{\scriptsize
\begin{equation}
    R_{\text{compliance}} = \sum_t R_{\text{compliance}}^t
\end{equation}
}
This mechanism reinforces the learning of structured interview processes and ensures the timely completion of a diagnosis, closely mirroring the practical constraints and expectations of real-world clinical environments.

By combining these three critical components, the total consultation evaluation score, or reward, received by the doctor agent at each time step \( t \) is calculated as:

{\scriptsize
\begin{equation}
    R = R_{\text{accuracy}} + R_{\text{information}} + R_{\text{compliance}}
\end{equation}
}
This sophisticated, multi-dimensional Consultation Evaluator not only guides the model toward superior diagnostic accuracy but also actively encourages the development of efficient information-gathering strategies and professional, compliant clinical behavior, ultimately aiming to achieve diagnostic capabilities that closely resemble those of expert human doctors.

% \subsubsection{Multi-agent Collaboration Mechanism}
% %图2:强化学习和多agent各模块协作设计
% \begin{figure*}[]
%     \centering
%     \includegraphics[width=1.0\textwidth]{framework.pdf} % 图片路径
%     \caption{The multi-agent collaborative reinforcement learning framework for \Ours{}. During the rollout stage, multi-turn interactions are conducted between the doctor agent and the patient agent.} % 标题
%     \label{fig:framework} % 标签（用于交叉引用）
% \end{figure*}

\subsection{Training Procedure}

% \begin{figure*}[]
%     \centering
%     \includegraphics[width=1.0\textwidth]{framework.pdf} % 图片路径
%     \caption{The multi-agent collaborative reinforcement learning framework for \Ours{}. During the rollout stage, multi-turn interactions are conducted between the doctor agent and the patient agent.} % 标题
%     \label{fig:framework} % 标签（用于交叉引用）
% \end{figure*}

As illustrated in Figure \ref{fig:framework}, our training framework for the doctor agent is built upon Qwen2.5-7B-Instruct, following the DeepSeek-R1 training paradigm \cite{guo2025deepseek}. The approach employs a two-stage training pipeline, integrating SFT and RL to cultivate clinical reasoning capabilities. Detailed experimental settings are illustrated in Supplementary Material.

Specifically, we randomly select 1,000 multi-turn consultation dialogues from the training corpus. Each doctor's query in the sampled dialogues is augmented with structured thought processes using DeepSeek-V3. These include hypothesis generation, evidence evaluation, and differential diagnosis steps derived from context. The doctor agent is fine-tuned on this enriched dataset to activate core capabilities in question-asking, diagnostic reasoning, and recommendation generation. 

Following SFT, we refine the agent's decision-making under interaction constraints using the policy optimization algorithm detailed in the Task Formulation section. To enhance robustness and mimic real-world clinical scenarios, we introduce a \textbf{Dynamic Turn Budget Training Strategy}. Each training episode is assigned a random dialogue turn budget (2--10 turns).
The model is explicitly reminded of the remaining turns after each interaction step, encouraging efficient information gathering. This two-stage approach ensures the agent first internalizes clinical reasoning patterns via supervised learning, then refines its strategy through interactive reward optimization. Ablation studies comparing alternative training strategies (direct SFT, direct RL, fixed-turn training) are presented in the experimental section.

\subsection{Dataset creation process}
% MTMedDialog is derived from three Chinese medical dialogue benchmarks: IMCS21 \cite{10.1093/bioinformatics/btac817}, CHIP-MDCFNPC \cite{zhang2021cblue}, and MedDG \cite{liu2022meddg}. As the source datasets were collected from real doctor-patient conversations, we implemented a two-stage denoising strategy: (1) filtering shallow dialogues with less than three turns through exact turn-count matching, and (2) removing noisy segments containing consecutive meaningless responses using DeepSeek-V3 \cite{liu2024deepseek}. We strictly preserved the original data partitioning protocol during this cleaning process to ensure evaluation reliability. The retained dialogues were then translated into English using DeepSeek-V3. Finally, we performed a manual inspection and refinement of the test set.

MTMedDialog is derived from three Chinese medical dialogue benchmarks: IMCS21 \cite{10.1093/bioinformatics/btac817}, CHIP-MDCFNPC \cite{zhang2021cblue}, and MedDG \cite{liu2022meddg}. As the source datasets were collected from real doctor-patient conversations, we implemented a two-stage denoising strategy: (1) filtering shallow dialogues with less than three turns through exact turn-count matching, and (2) removing noisy segments containing consecutive meaningless responses using DeepSeek-V3 \cite{liu2024deepseek}. We strictly preserved the original data partitioning protocol during this cleaning process to ensure evaluation reliability. 
The retained dialogues were then translated into English using DeepSeek-V3 to enable fair comparison with international frontier models (GPT-4o, DeepSeek-V3, LLaMA-3.1), which are predominantly trained on English corpora, and to create the first English dataset supporting dynamic patient simulation. The test set underwent manual inspection and refinement by bilingual clinical experts to ensure quality.

The test set serves dual evaluation purposes: 1) assessing doctor agent's questioning and diagnostic capabilities using complete dialogue trajectories with gold-standard diagnosis labels; 2) evaluating patient agent's response quality through a randomly selected subset of 500 samples. We will release both Chinese and English versions of the test set, allowing researchers to choose based on their needs.

\backmatter

\section{Data Availability}
The codes and datasets are available for access at 
\href{https://github.com/JarvisUSTC/DoctorAgent-RL.git}{https://github.com/JarvisUSTC/DoctorAgent-RL.git}. 

\section{Competing Interests}
The authors declare no competing interests.

\section{Funding}
This work was supported in part by funds from the Major Project of Guangzhou National Laboratory (No. SRPG22-007), the Major Project of Guangzhou National Laboratory (No. GZNL2025C01013), the Guangdong Basic and Applied Basic Research Foundation (No. 2025A1515011597), and the Prevention and Control of Emerging and Major Infectious Diseases-National Science and Technology Major Project (No. 2025ZD01901900).
% This work was supported in part by funds from the Major Project of Guangzhou National Laboratory (No. SRPG22007); the Major Project of Guangzhou National Laboratory (No. GZNL2025C01013); the Guangdong Basic and Applied Basic Research Foundation (No.2025A1515011597).

\section{Acknowledgments}
We acknowledge the technical support from the Data Science Platform of Guangzhou National Laboratory and Bio-medical big data Operating System (Bio-OS), as well as the support from the InnoHK Program.

\section{Author contributions}
Conceptualization, Y.F. and J.W.; methodology, Y.F. and J.W.; formal analysis, Y.F.; investigation, Y.F. and J.W.; resources, Y.L.; data curation, Y.F. and J.W.; writing—original draft preparation, Y.F. and J.W.; writing—review and editing, Y.F., J.W., L.Z, Y.Z, Z.L. and Y.L.; visualization, Y.F.; supervision, Y.L. and Z.L.; All authors have read and agreed to the published version of the manuscript.

\section{Additional information}
\subsection{Supplementary information}
Supplementary information is available at Supplementary material.pdf.
\subsection{Corresponding authors}
Correspondence to \href{mailto:zhen.lei@ia.ac.cn}{Zhen Lei}, \href{mailto:li_yixue@gzlab.ac.cn}{Yixue Li}.

% \begin{appendices}

% \section{Section title of first appendix}\label{secA1}

% An appendix contains supplementary information that is not an essential part of the text itself but which may be helpful in providing a more comprehensive understanding of the research problem or it is information that is too cumbersome to be included in the body of the paper.




% \end{appendices}


\bibliography{sn-bibliography}% common bib file

\end{document}

\title{Real-World Doctor Agent with Proactive Consultation through Multi-Agent Reinforcement Learning}


\author[1,2,3]{\fnm{Yichun} \sur{Feng}\orcidlink{0009-0007-4511-4713}}\email{fengyichun22@mails.ucas.ac.cn}\equalcont{These authors contributed equally to this work.}

\author[4]{\fnm{Jiawei} \sur{Wang}\orcidlink{0009-0005-0575-2695}}\email{wangjiawei@mail.ustc.edu.cn}
\equalcont{These authors contributed equally to this work.}

\author[3]{\fnm{Lu} \sur{Zhou}\orcidlink{0009-0003-6538-0649}}\email{zhou\_lu@gzlab.ac.cn}
\author[5]{\fnm{Yikai} \sur{Zheng}\orcidlink{0009-0004-8709-4536}}\email{zhengyk23@mail2.sysu.edu.cn}
\author*[1,2,9]{\fnm{Zhen} \sur{Lei}\orcidlink{0000-0002-0791-189X}}\email{zhen.lei@ia.ac.cn}
\author*[3,6,7,8]{\fnm{Yixue} \sur{Li}\orcidlink{0000-0002-1198-7176}}\email{li\_yixue@gzlab.ac.cn}

\affil[1]{\orgdiv{School of Advanced Interdisciplinary Sciences}, \orgname{University of Chinese Academy of Sciences}, \orgaddress{\street{380, Huaibei Town, Huairou District}, \city{Beijing}, \postcode{100049}, \state{Beijing}, \country{China}}}

\affil[2]{\orgdiv{Institute of Automation}, \orgname{Chinese Academy of Sciences}, \orgaddress{\street{No. 95, Zhongguancun East Road}, \city{Beijing}, \postcode{100190}, \country{China}}}

\affil[3]{\orgname{Guangzhou National Laboratory}, \orgaddress{\street{No. 9 XingDaoHuanBei Road, Guangzhou International Bio Island}, \city{Guangzhou}, \postcode{510005}, \state{Guangdong}, \country{China}}}

\affil[4]{\orgdiv{Department of EEIS}, \orgname{University of Science and Technology of China}, \orgaddress{\street{No. 443, Huangshan Road, Shushan District}, \city{Hefei}, \postcode{230026}, \state{Anhui}, \country{China}}}
\affil[5]{\orgdiv{School of Intelligent Systems Engineering}, \orgname{Sun Yat-sen University}, \orgaddress{\street{No. 66, Gongchang Road, Guangming District}, \city{Shenzhen}, \postcode{518107}, \state{Guangdong}, \country{China}}}
\affil[6]{\orgdiv{GMU-GIBH Joint School of Life Sciences}, \orgname{The Guangdong-Hong Kong-Macau Joint Laboratory for Cell Fate Regulation and Diseases}, \orgname{Guangzhou Medical University}, \orgaddress{\city{Guangzhou}, \postcode{511436}, \country{China}}}

\affil[7]{\orgdiv{School of Life Sciences and Biotechnology}, \orgname{Shanghai Jiao Tong University}, \orgaddress{\city{Shanghai}, \postcode{200240}, \country{China}}}

\affil[8]{\orgdiv{Shanghai Institute of Nutrition and Health}, \orgname{Chinese Academy of Sciences}, \orgaddress{\city{Shanghai}, \postcode{200030}, \country{China}}}

% \affil[9]{\orgdiv{Bioland Laboratory}, \orgaddress{\city{Guangzhou}, \postcode{510005}, \country{China}}}
\affil[9]{\orgdiv{Centre for Artificial Intelligence and Robotics, Hong Kong Institute of Science \& Innovation}, \orgname{Chinese Academy of Sciences}, \orgaddress{\city{Hong Kong}, \country{China}}}
\maketitle
% \section{Details of MTMedDialog}
% \label{sec:dataset_details}

% This section presents the statistical distribution of disease categories in the MTMedDialog test set. We employed the DeepSeek-V3 to automatically classify the diagnostic results of each data entry, strictly adhering to a predefined eight-category disease classification system. Samples that did not conform to this classification framework were subsequently removed, resulting in a final collection of 2,082 high-quality test samples. Detailed data are shown in Table~\ref{tab:app_disease_dist}.

% \begin{table*}[ht]
% \centering
% \begin{tabular}{lc}
% \toprule
% \textbf{Disease Category} & \textbf{Number of Cases} \\
% \hline
% Digestive System Diseases & 1,290 \\
% Respiratory System Diseases & 402 \\
% Infectious Diseases & 118 \\
% Genitourinary System Diseases & 100 \\
% Neurological Disorders & 77 \\
% Circulatory System Diseases & 48 \\
% Endocrine Disorders & 27 \\
% Skin Diseases & 20 \\\hline
% % \bottomrule
% \end{tabular}
% \caption{Disease Category Distribution in MTMedDialog Test Set}
% \label{tab:app_disease_dist}
% \end{table*}

\section{Prompts for Evaluation Metrics}
\label{sec:Evaluation_Metrics}
This section will provide a detailed explanation of the evaluation prompts for both the doctor agent and patient agent. The evaluation prompt for the doctor agent's Diagnosis and Recommendation Accuracy is presented in Figure~\ref{fig:evl_doctor}, while the evaluation prompt for the patient agent is shown in Figure~\ref{fig:evl_patient}.

\begin{figure*}[ht]
  \centering
    \begin{tikzpicture}
      % Main box with light gray background
      \draw[thick,rounded corners=10pt,fill=gray!10] (-6.5,-0.5) rectangle (6.5,8.0);
    
      % Title box - black, full width, top-aligned
      \filldraw[black,rounded corners=5pt] (-6.5,8.0) rectangle (6.5,7.4)
        node[midway,white] {Prompt for Evaluating the Doctor Agent's Diagnosis and Recommendation Accuracy};
    
      % Text inside the box
      \node[align=left, text width=12cm] at (0,3.5) {
 Task: As a medical expert, evaluate the semantic similarity between the model-generated medical text and the ground truth reference. Score on a 0–5 point scale based on meaning alignment (wording differences are acceptable if meaning matches).\\
Criteria:\\
5: Identical meaning (different wording okay).\\
4: Minor wording/detail differences; overall meaning aligned.\\
3: Partial meaning overlap; important differences exist but core intent is partially shared.\\
2: Limited meaning overlap; key details or context differ significantly.\\
1: Minimal meaning overlap; mostly unrelated or only superficially related.\\
0: Unrelated or completely different meaning (no meaningful semantic connection).

Now evaluate:\\
Candidate: "\{candidate\}"\\
Reference: "\{reference\}"\\
% RESPONSE:
OUTPUT FORMAT: \texttt{<think>} [Your Thinking Process] \texttt{</think>}\texttt{<answer>} [Your Score] \texttt{</answer>}
      };
    \end{tikzpicture}
  \caption{Prompt for Evaluating the Doctor Agent's Diagnosis and Recommendation Accuracy. \{candidate\} represents the model-generated medical text (doctor agent's output), while \{reference\} denotes the ground truth clinical reference standard.}
  \label{fig:evl_doctor}
\end{figure*}

\begin{figure*}[ht]
  \centering
    \begin{tikzpicture}
      % Main box with light gray background
      \draw[thick,rounded corners=10pt,fill=gray!10] (-6.5,-0.5) rectangle (6.5,16.0);
    
      % Title box - black, full width, top-aligned
      \filldraw[black,rounded corners=5pt] (-6.5,16.0) rectangle (6.5,15.4)
        node[midway,white] {Prompt for Evaluating the Patient Agent};
    
      % Text inside the box
      \node[align=left, text width=12cm] at (0,7.5) {
      You are evaluating the quality of a patient's simulated responses during a medical consultation.\\[6pt]
      
      \textbf{Original self-report:} \{self\_report\}\\[3pt]
      \textbf{Diagnosis:} \{diagnosis\}\\[3pt]
      \textbf{Recommendation:} \{recommendation\}\\[3pt]
      \textbf{Doctor's questions and simulated patient answers:} \{simulated\_dialogue\}\\[6pt]
      
      \textbf{Instructions:}\\[3pt]
      1. \textbf{Information Control Rate (0-1):}\\
      \quad - Check whether the patient's answers include extra information not asked by the doctor.\\
      \quad - Deduct 20\% for each extra information point.\\
      \quad - Score = max(0, 1 - 0.2 $\times$ number\_of\_extra\_points)\\[6pt]
      
      2. \textbf{Response Completeness Rate (0-1):}\\
      \quad - Check whether the patient answered all points asked by the doctor.\\
      \quad - Deduct 20\% for each missing information point.\\
      \quad - Score = max(0, 1 - 0.2 $\times$ number\_of\_missing\_points)\\[6pt]
      
      3. \textbf{Factual Conflict Rate (0-1):}\\
      \quad - Check if the patient's response is completely opposite to the original self-report.\\
      \quad - Increase 20\% for each completely opposite found.\\
      \quad - Score = min(1, 1 - 0.2 $\times$ number\_of\_opposite)\\[6pt]
      
      \textbf{Return ONLY in JSON format:}\\
      \{\\
      \quad "information\_control\_rate": float,\\
      \quad "response\_completeness\_rate": float,\\
      \quad "factual\_conflict\_rate": float\\
      \}
      };
    \end{tikzpicture}
  \caption{Prompt for Evaluating the Patient Agent.}
  \label{fig:evl_patient}
\end{figure*}

\section{Prompt for Doctor Agent}
\label{sec:pro_doctor}
This section elaborates on the prompt design for the doctor agent during both training and inference phases. The complete prompt structure is illustrated in Figure~\ref{fig:pro_doctor}.
\begin{figure*}[ht]
  \centering
    \begin{tikzpicture}
      % Main box with light gray background
      \draw[thick,rounded corners=10pt,fill=gray!10] (-6.5,-0.5) rectangle (6.5,20.0);
    
      % Title box - black, full width, top-aligned
      \filldraw[black,rounded corners=5pt] (-6.5,20.0) rectangle (6.5,19.4)
        node[midway,white] {Prompt for Doctor Agent};
    
      % Text inside the box
      \node[align=left, text width=12cm] at (0,9.5) {
 You are an experienced doctor who needs to provide professional diagnosis and advice to patients through consultation. Please listen carefully to the patient's description, ask targeted questions, and collect sufficient information before giving a diagnosis and treatment recommendation.\\[6pt]
      
      \textbf{Quick Guide}\\[3pt]
      \textbf{Objectives:}\\[3pt]
      1. Obtain key information through effective questioning, each round of questions should be modified based on the previous round's content, meaning you shouldn't ask similar questions.\\
      2. Comprehensively analyze the patient's condition to provide an accurate diagnosis and appropriate treatment recommendations.\\[6pt]
      
      \textbf{Rules:}\\[3pt]
      1. You can only choose one of the options to respond, you cannot both answer questions and provide a diagnosis simultaneously.\\
      2. Absolutely do not repeat or ask questions similar or identical to those previously asked.\\[6pt]
      
      \textbf{Response:}\\[3pt]
      \texttt{<think>} [your thinking] \texttt{</think>}\\
      \texttt{<answer>}If you believe there is insufficient information, please only ask one question, in this format:\\
      Question: (your question).\\
      \texttt{</answer>} | \texttt{<answer>}If you believe you have obtained enough information, please only provide diagnosis and recommendations, in this format:\\
      Diagnosis: (the patient's most likely disease or symptoms)\\
      Recommendation: (corresponding treatment plan or advice)\\
      \texttt{</answer>}\\[6pt]
      
      \textbf{Rewards:}\\[3pt]
      Incorrect format: -2.0\\
      Effective question (patient can provide an answer and the question is helpful for diagnosis): +1.0\\
      Ineffective questions do not count towards score\\
      Repeated questions: -2.0\\
      The number of conversation turn is limited. Reaching maximum interaction rounds without providing a diagnosis: -5.0\\
      Completely correct diagnosis and recommendations: +10.0
      };
    \end{tikzpicture}
  \caption{Prompt for Doctor Agent.}
  \label{fig:pro_doctor}
\end{figure*}

\section{Prompt for Patient Agent}
\label{sec:pro_patient}
This section first details the patient agent's implicit disease knowledge generation mechanism, with the complete prompt structure shown in Figure~\ref{fig:enhance_patient}. Subsequently, it examines the agent's prompt design methodology for both training and inference phases, as fully illustrated in Figure~\ref{fig:pro_patient}.
\begin{figure*}[ht]
  \centering
    \begin{tikzpicture}
      % Main box with light gray background
      \draw[thick,rounded corners=10pt,fill=gray!10] (-6.5,-0.5) rectangle (6.5,8.0);
    
      % Title box - black, full width, top-aligned
      \filldraw[black,rounded corners=5pt] (-6.5,8.0) rectangle (6.5,7.4)
        node[midway,white] {Prompt for Patient Agent to Develop a Comprehensive Implicit Health Profile};
    
      % Text inside the box
      \node[align=left, text width=12cm] at (0,3.5) {
 As a medical assistant, expand the patient's symptom description based on:\\
        \begin{itemize}
            \item Original self-report: \texttt{\{self\_report\}}
            \item Dialogue history: \texttt{\{dialogue\_history\}}
            \item Diagnosis: \texttt{\{diagnosis\}}
            \item Recommendation: \texttt{\{recommendation\}}
        \end{itemize}
        
        \textbf{Processing Rules:}
        \begin{enumerate}
            \item Summarize the patient's information: Combine the 'Original self-report' and all patient responses from 'dialogue' into a single coherent paragraph. Include only factual patient statements and exclude the doctor's questions. If a patient response only makes sense in the context of the doctor's question, infer its meaning based on the context.
            
            \item Based on diagnosis and recommendations, add medical evidence to clearly support symptoms.
            
            \item Never contradict the patient's original statements.
            
            \item Keep the language natural and clinical.
             \item Return ONLY the enhanced description.
        \end{enumerate}

      };
    \end{tikzpicture}
  \caption{Prompt for Patient Agent to Develop a Comprehensive Implicit Health Profile.}
  \label{fig:enhance_patient}
\end{figure*}

\begin{figure*}[ht]
  \centering
    \begin{tikzpicture}
      % Main box with light gray background
      \draw[thick,rounded corners=10pt,fill=gray!10] (-6.5,-0.5) rectangle (6.5,8.0);
    
      % Title box - black, full width, top-aligned
      \filldraw[black,rounded corners=5pt] (-6.5,8.0) rectangle (6.5,7.4)
        node[midway,white] {Prompt for Patient Agent Training and Inference};
    
      % Text inside the box
      \node[align=left, text width=12cm] at (0,3.5) {
    You are interacting with a doctor.\\ \textbf{Medical Response Instructions:}\\
     Answer each medical question concisely in one sentence, strictly describing symptoms while avoiding any mention of diagnoses or recommendations.\\
      If the question is unrelated to your chief complaint, state: "Sorry, I cannot answer this question."\\
      If the question is repetitive, reply: "Sorry, you've already asked this question."\\
      \textbf{Your chief complaint: }\\
      \{description\}\\
      \textbf{doctor's question history:}\\
        \{history\_questions\}\\
      \textbf{Current doctor question: }\\
      \{question\}\\
      \textbf{Output format:}\\
      \texttt{<think>}[Your reasoning]\texttt{</think>}\texttt{<answer>}[Your response]\texttt{</answer>}

      };
    \end{tikzpicture}
  \caption{Prompt for Patient Agent Training and Inference.}
  \label{fig:pro_patient}
\end{figure*}

\section{Detailed Description of the Dialogue Flow in \Ours{}}
\label{sec:example}

\begin{figure*}[htbp]
    \centering
    \includegraphics[width=0.8\textwidth]{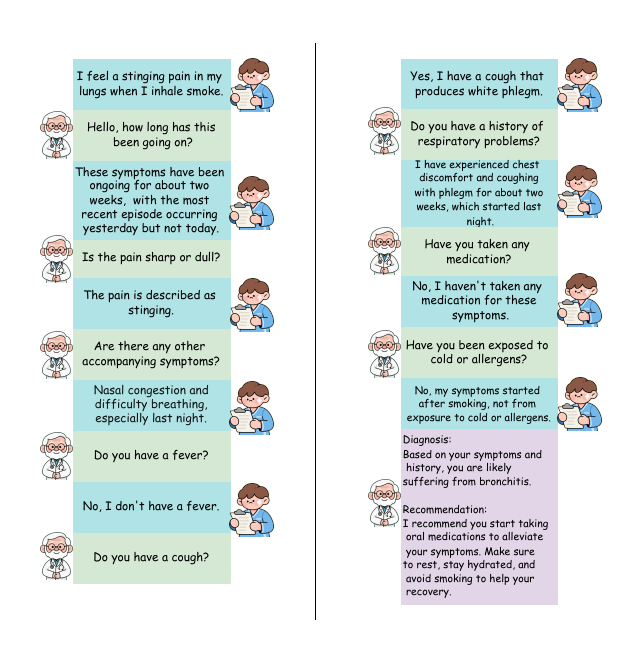} % 图片路径
    \caption{An example of dialogue flow among Doctor and Patient in \Ours{}.} % 标题
    \label{fig:example} % 标签（用于交叉引用）
\end{figure*}

This section provides a detailed explanation of the dialogue flow in \Ours{}, as illustrated in Figure ~\ref{fig:example}. The framework aims to simulate a realistic diagnostic process by coordinating a structured dialogue flow between two agents: the doctor and the patient. This appendix comprehensively describes the mechanisms for dialogue initiation, progression, and termination.  

The diagnostic process unfolds through multiple rounds of interaction between the two agents. The patient's self-report serves as the starting point for the first round of dialogue between the doctor and patient agents. During this phase, the doctor agent actively conducts a comprehensive inquiry, collecting diagnostic information through targeted questioning. The patient agent, functioning as a non-player character (NPC), provides feedback on their condition to the doctor agent in each dialogue round based on carefully designed prompts.  

The diagnostic phase concludes when either of the following conditions is met: (1) The doctor agent determines that the collected information is sufficient for diagnosis and directly outputs the diagnostic results. (2) The predefined maximum number of interaction rounds is reached, compelling the doctor agent to provide a final diagnosis. This mechanism ensures a structured and finite diagnostic process.

\section{Settings of RL training in \Ours{}}
\label{appendix:settings}
\textbf{Hardware \& Software.} All training and evaluation were performed on a system featuring eight NVIDIA A100 80GB PCIe GPUs, an Intel Xeon Platinum 8369B 32-Core Processor, and 1.0 TB of RAM. For supervised fine-tuning, we utilized the LLaMA-Factory framework \cite{zheng2024llamafactory} to fine-tune with LoRA \cite{hu2022lora}. Our \Ours{}, is built upon the VERL framework (v0.2) \cite{sheng2024hybridflow} for reinforcement learning with language models, with RAGEN \cite{wang2025ragen} providing the multi-turn RL architecture. We employed vLLM (v0.8.5) \cite{kwon2023efficient} for efficient LLM inference and evaluation, PyTorch (v2.4.0) with CUDA 12.4 for deep learning, and Ray for distributed training and serving. Flash Attention 2 \cite{dao2023flashattention} was integrated to optimize attention computation.

\textbf{Hyperparameter.} For completeness and reproducibility, all hyperparameters employed in \Ours{} are detailed in Table \ref{tab:hyperparameters}. We observed that the decoupled clip approach from DAPO \cite{yu2025dapo} significantly enhances exploration during RL training, and thus, we adopted it for our final training process.

\begin{table*}[ht]
    \centering
    % \caption{Hyperparameters for Supervised Fine-Tuning and Reinforcement Learning experiments.} % Adjusted caption
    % \label{tab:hyperparameters} % New label
    \begin{tabular}{llc} % Changed to llcl as one data column is removed, and now there are three columns
        \toprule
        \multirow{3}{*}{\shortstack{Supervised \\ Fine-Tuning}} & learning rate & 1e-4 \\
        & batch size & 64 \\
        & epochs & 3 \\
        & lora\_rank & 8 \\
        \midrule
        \multirow{10}{*}{\shortstack{\Ours{}}} & actor learning rate & 1e-6 \\
        & state masking & true \\
        & kl loss coef & 0.001 \\
        & kl penalty & low\_var\_kl \\
        & entropy coeff & 0.001 \\
        & clip high & 0.28 \\
        & clip low & 0.2 \\
        & batch size & 128 \\
        & epochs & 1 \\
        & rollout group size & 8 \\
        & rollout temperature & 0.7 \\
        \bottomrule
    \end{tabular}
        \caption{Hyperparameters for Supervised Fine-Tuning and Reinforcement Learning experiments.} % Adjusted caption
    \label{tab:hyperparameters} % New label
\end{table*}

\clearpage
\bibliography{sn-bibliography}% common bib file